\pdfoutput=1

\documentclass[11pt]{article}

\usepackage{acl}

\usepackage{times}
\usepackage{latexsym}

\usepackage[T1]{fontenc}

\usepackage[utf8]{inputenc}

\usepackage{microtype}

\usepackage{inconsolata}

\usepackage{caption}
\usepackage{subcaption}
\usepackage{amsmath,wasysym,amsfonts}
\usepackage[export]{adjustbox}
\usepackage{CJKutf8}
\usepackage{enumitem}
\usepackage{multirow}

\DeclareMathOperator*{\argmin}{arg\,min}

\newcommand{\ku}{$^{\heartsuit}$}
\newcommand{\eth}{$^{\spadesuit}$}

\title{Understanding Subword Compositionality of Large Language Models}

\author{Qiwei Peng\ku,~Yekun Chai\eth,~Anders Søgaard\ku~ \\
{\ku}University of Copenhagen \quad {\eth}ETH Zurich \\
\texttt{\{qipe,soegaard\}@di.ku.dk}\quad \texttt{yechai@ethz.ch}
}

\begin{document}
\maketitle
\begin{abstract}


Large language models (LLMs) take sequences of subwords as input, requiring them to effective compose subword representations into meaningful word-level representations. In this paper, we present a comprehensive set of experiments to probe how LLMs compose subword information, focusing on three key aspects: structural similarity, semantic decomposability, and form retention. 
Our analysis of the experiments suggests that five LLM families can be classified into three distinct groups, likely reflecting difference in their underlying composition strategies. Specifically, we observe (i) three distinct patterns in the evolution of structural similarity between subword compositions and whole-word representations across layers; (ii) great performance when probing layer by layer their sensitivity to semantic decompositionality; and (iii) three distinct patterns when probing sensitivity to formal features, e.g., character sequence length. These findings  provide valuable insights into the compositional dynamics of LLMs and highlight different compositional pattens in how LLMs encode and integrate subword information.

\end{abstract}  

\section{Introduction}
Large language models (LLMs) rely heavily on subword tokenization \citep{achiam2023gpt, dubey2024llama} that processes words into a sequence of subwords which potentially disrupts morpheme boundaries \citep{batsuren2024evaluating}. Despite this, LLMs have demonstrated impressive capability in comprehending word meanings \citep{shani-etal-2023-towards, xu2024tip}, suggesting that they effectively construct meaningful word representations from subword components. One possible approach to this is memorization, where models store entire input-output pairs. This strategy, adopted by Ned Block’s {\em humongous table program} \citep{Block1981-BLOPAB}, scales only if all input-output pairs have been seen during training. However, this is computationally infeasible due to the exponential growth in possible combinations with increasing input length and vocabulary size. Given their promising ability on word meaning understanding, LLMs must be employing systematic compositional strategies rather than relying solely on memorization to generalize beyond seen data. This motivates our investigation into how LLMs construct word representations from subword components and uncover potential consistent and systematic patterns in subword composition.

To systematically examine these compositional strategies, we analyze subword composition from three key perspectives. First, we examine how the geometry of composed word representations relates to that of their subword constituents. Specifically, we assess whether composed representations maintain \textit{linear alignment} with their constituent representations, revealing patterns of structural similarity across layers. Prior studies have explored geometry properties of word and phrase embeddings they construct \citep{gong2017geometry}, and examined distances between composed subwords and full-word embeddings in vector space \citep{chai-etal-2024-tokenization}. Our focus here is to identify linear alignment patterns that reveal structural similarity and transformation dynamics between composed representations and whole-word representations across layers.

Second, we probe whether composed representations encode fundamental aspects of word meaning, particularly the distinction between semantically decomposable and non-decomposable words. Building on previous work that assessed embeddings for their awareness of syntactic and semantic properties, such as sentence length, tense, and identification of semantic roles \citep{conneau-etal-2018-cram, ettinger-etal-2018-assessing, klafka-ettinger-2020-spying}, our analysis focuses on whether LLMs preserve relevant information on semantic decompositionality during composition. Third, we investigate whether composed representations retain surface-level features, such as word length across models and layers. While some models exhibit strong retention of such features, others abstract away form-related information, which shows variations in how form and content are preserved. 
By analyzing LLMs across these dimensions, our study provides valuable insights into how LLMs process subwords and form word-level representations, contributing to a broader understanding of compositional dynamics in LLMs.

\paragraph{Contributions} In this work, we present a set of new experiments designed to probe the compositional dynamics of LLMs around subwords. Our experiments with six different LLMs across five LLM families on three types of tasks demonstrate that: 
    (i) In most models, subword composition is {\bf isometric to simple addition}.
    (ii) Content information such as semantic decompositionality is well-preserved in the composed representation for all models across all layers. Formal information about word length, in contrast, is only preserved in some models. This has direct implications for the {\bf derivability of form and content} of the input. 
    (iii) The six LLMs fall into three groups, relying on {\bf three distinct compositional strategies}, i.e., ways of constructing composed representations from subwords.


\section{Related Work} 
\paragraph{Tokenization} Current generations of LLMs  \citep{achiam2023gpt, touvron2023llama, team2023gemini, lozhkov2024starcoder}, heavily rely on subword tokenization where an input text is split into a sequence of subwords derived from a predefined vocabulary. Such approaches include frequency-based methods such as Byte-Pair Encoding \citep{sennrich-etal-2016-neural} and Byte-level BPE \citep{wang2020neural}, probability-based methods such as WordPiece \citep{schuster2012japanese} and Unigram \citep{kudo-2018-subword}. Tokenization approaches need to balance the trade-off between vocabulary size and diverse language coverage in multilingual scenarios. Tokenization-free or pixel-based approaches have been proposed to side-step this trade-off \citep{rust2023language, tai-etal-2024-pixar, chai-etal-2024-autoregressive}, and various tasks have been proposed to better examine the impact and robustness of subword tokenization \citep{gee-etal-2022-fast, cao-etal-2023-unnatural, chai-etal-2024-tokenization, wang2024tokenization, batsuren2024evaluating}. Our work aims to understand subword compositionality in LLMs.

\paragraph{Compositionality} The compositional ability allows models to generalize beyond simple memorization. Previous works have thoroughly examined compositionality in phrase \citep{yu-ettinger-2020-assessing, bertolini-etal-2021-representing} and sentence embeddings \citep{dasgupta2018evaluating, xu2023compositionality}. Recent studies have also explored general compositional behaviors of LLMs in reasoning tasks \citep{dziri2024faith, li-etal-2024-understanding} and rule following \citep{wang-etal-2024-llms}. Our work sets out to investigate whether subwords, as a result of tokenization, exhibit any compositional dynamics through geometry and probing analysis. Procrustes analysis, which is a form of statistical shape analysis \citep{schonemann1966generalized}, is widely used to analyze structural similarity between two language spaces \citep{peng-sogaard-2024-concept} and modality spaces \citep{li2024vision}. Additionally,  probing analysis is a standard approach for dissecting syntactic and semantic features in neural models, such as syntactic depth, tense, and semantic roles \citep{ettinger-etal-2018-assessing, conneau-etal-2018-cram, hewitt-manning-2019-structural, klafka-ettinger-2020-spying}. 

\begin{figure*}[!ht]
    \centering
    \includegraphics[width=0.85\textwidth]{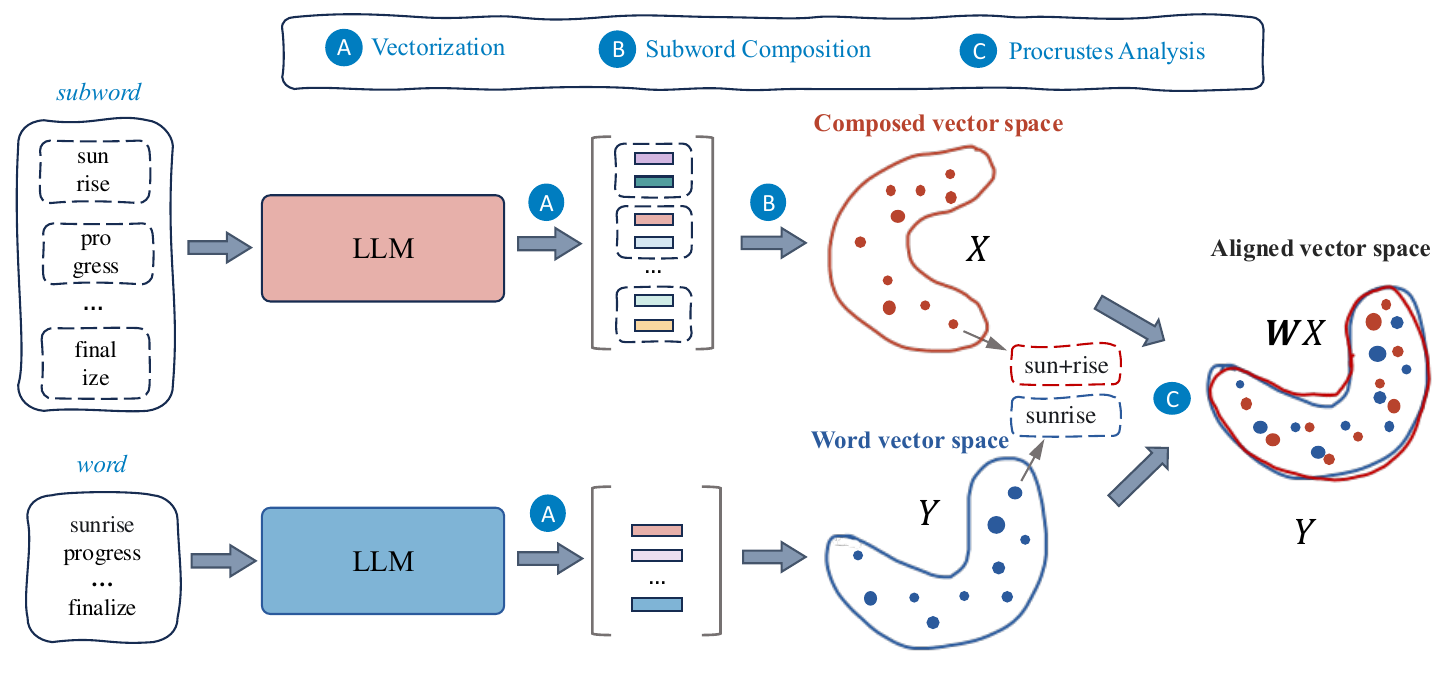}
     \vspace{-1em}
    \caption{Illustration of the pipeline of our geometry analysis. All words and subwords exist in models' vocabulary. Vector representations are first obtained by feeding them into LLMs. Composed vector space is then constructed by applying composition operations among subword representations. Procrustes analysis is performed between the original word vector space and the composed vector space to find the linear alignment.}
    \label{fig:geometry_analysis_method} 
    \vspace{-10pt}
\end{figure*}

\section{Geometry Analysis}
We first conduct geometry analysis on the internal vector space of different LLM, focusing on the structural similarity between composed representations and the original whole word representation.     

\subsection{Dataset}
\label{sec:dataset}
\citet{batsuren-etal-2022-sigmorphon} proposed a benchmark on morpheme segmentation which collected more than 577,374 unique English words with its morphological categories. We take advantage of this resource and pick out words that have both its whole word form and potential subwords in the model's vocabulary. In this work, we specifically focus on two-subword combination (e.g., limit $\Rightarrow$ (li, mit)\footnote{\textit{limit}, \textit{li}, and \textit{mit} are all in model's vocabulary.}). After going through six different language models, we end up with a parallel\footnote{It is parallel in the whole word form, while the tokenized results might be different.} dataset across these language models. In total, we have 3,432 words covering 2,316 root words (words that are free morphemes, such as \textit{dog} and \textit{progress}) and 1,116 non-root words (words that fall into other morphological categories such as inflection only, e.g., \textit{prepared}, derivation only, e.g., \textit{intensive}, and compound, e.g., \textit{hotpot}). As one word could have multiple subword combinations discovered (e.g., numeric $\Rightarrow$ (n, umeric), (num, eric), (numer, ic)), we have all combinations included. In the following experiments, we conduct 3 runs where each run with randomly picked combination to reflect the variation. The ribbons in the experiment figures demonstrate standard deviations brought in by such variation. We randomly split these words into train, test splits.          

\begin{table}[!ht]
\centering
\resizebox{0.65\columnwidth}{!}{%
\begin{tabular}{|c|c|c|c|} 
\hline
      & Root & Non-Root & Total \\ \hline    
Train & 1852 & 893      & 2745  \\ \hline    
Test  & 464  & 223      & 687   \\ \hline   
Total & 2316 & 1116     & 3432  \\ \hline   
\end{tabular}%
}  
\caption{The statistics of the dataset.}
\label{tab:dataset} 
\vspace{-10pt}
\end{table}      

\paragraph{LLMs and Vector Representation} The six instruction-tuned LLMs we experiment include Llama3-8B-Instruct, Llama3.1-8B-Instruct \citep{dubey2024llama}, Aya-expanse-8B \citep{dang2024ayaexpansecombiningresearch}, Gemma2-9B-it \citep{gemma_2024}, Qwen2.5-7B-Instruct \citep{team2024qwen2}, and Falcon-7B-Instruct \citep{falcon40b}. All above models adopt subword tokenization strategy and are instruction-tuned. The whole word vector representation is derived through feeding the exact word to the model. Subword representations are obtained separately through the same pipeline. As all words and subwords exist in models' vocabulary, we can directly obtain their vector representations without additional operations. Different composition operations are then performed on subword representations to obtain the composed representation, which will later be compared against the original whole-word representation to examine structural similarity.

\subsection{Methods}
We utilize Procrustes Analysis \citep{schonemann1966generalized}, i.e., the induction of a linear projection between two subspaces, to quantify the {\em isometry} or structural similarity between whole word representations and composed representations of subwords. Assume $X$ and $Y$ are two matrices of size $n \times d$ ($n$ is the number of examples, and $d$ refers to the embedding dimension). Such that the $i$-th row of $X$ is the composed embedding of two subwords, and $i$th row of $Y$ is the original embedding of the whole word. The linear transformation is derived through singular value decomposition (SVD) of $YX^{T}$: 
\begingroup
\setlength{\abovedisplayskip}{8pt}
\setlength{\belowdisplayskip}{8pt} 
\begin{equation}
    W^{*} = \argmin_{W \in O_{d(\mathbb{R})}}||WX - Y||_{F} = UV^{T}
\end{equation}
\endgroup
where $U \Sigma V^{T} = \text{SVD}(YX^{T})$. With the obtained $W^{*}$, we transform composed embeddings $X$ into the original vector space. We then perform cosine similarity to retrieve the most similar original word vector. Following previous works on measuring representation alignment \citep{li2024vision,wu-etal-2024-representational}, we use Precision@1 (P@1) as our performance metric. The overall pipeline of the method is illustrated in Figure \ref{fig:geometry_analysis_method}. The train split is used to find the optimal linear transformation $W^{*}$ which will then be applied to the test split for evaluation.

\subsection{Results}
\paragraph{Main Geometry Results}
Our first experiment simply evaluates the structural similarity of LLM whole word representations and addition of, multiplication of, and absolute difference between constituent representations, by measuring their performance (P@1) across layers.

\begin{figure}[!ht]
     \centering
     \vspace{-8pt}
     \begin{adjustbox}{minipage=\columnwidth,scale=1} 
     \begin{subfigure}[t]{0.49\columnwidth}
         \centering
         \includegraphics[width=\columnwidth]{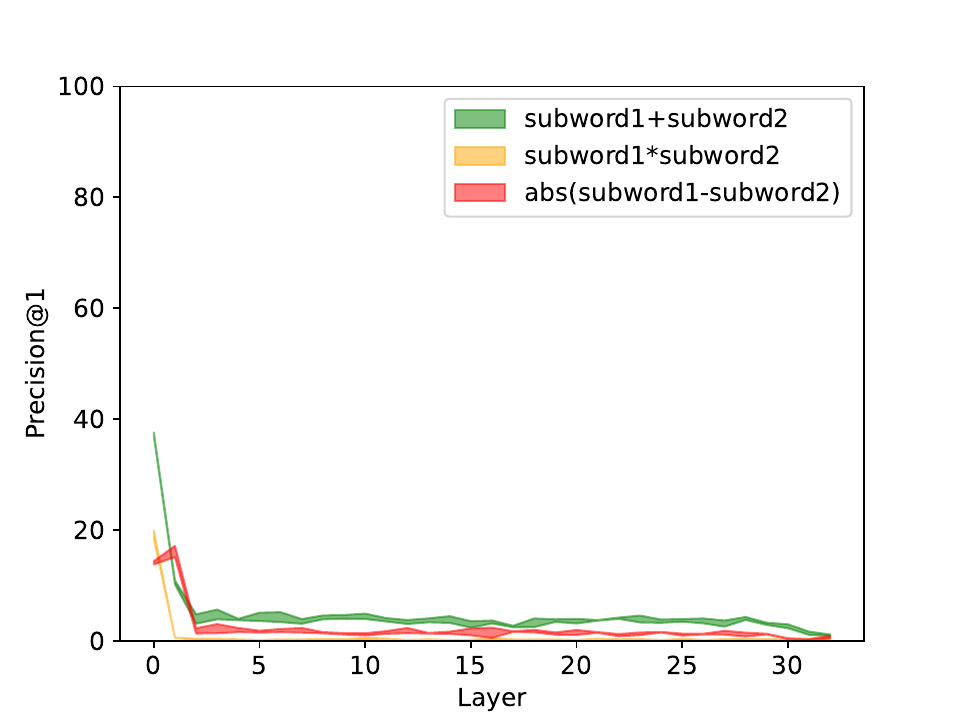}
         \caption{Llama3}
    \end{subfigure}
    \hfill
    \begin{subfigure}[t]{0.49\columnwidth}
         \centering
         \includegraphics[width=\columnwidth]{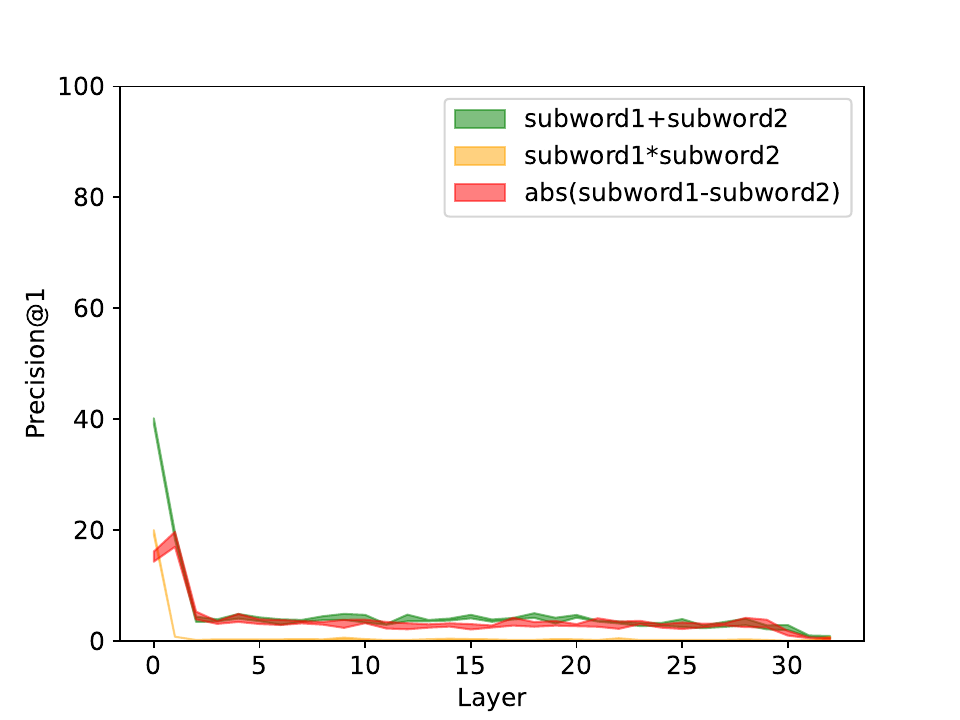}
         \caption{Llama3.1}
    \end{subfigure}
    \newline
     \begin{subfigure}[t]{0.49\columnwidth}
         \centering
         \includegraphics[width=\columnwidth]{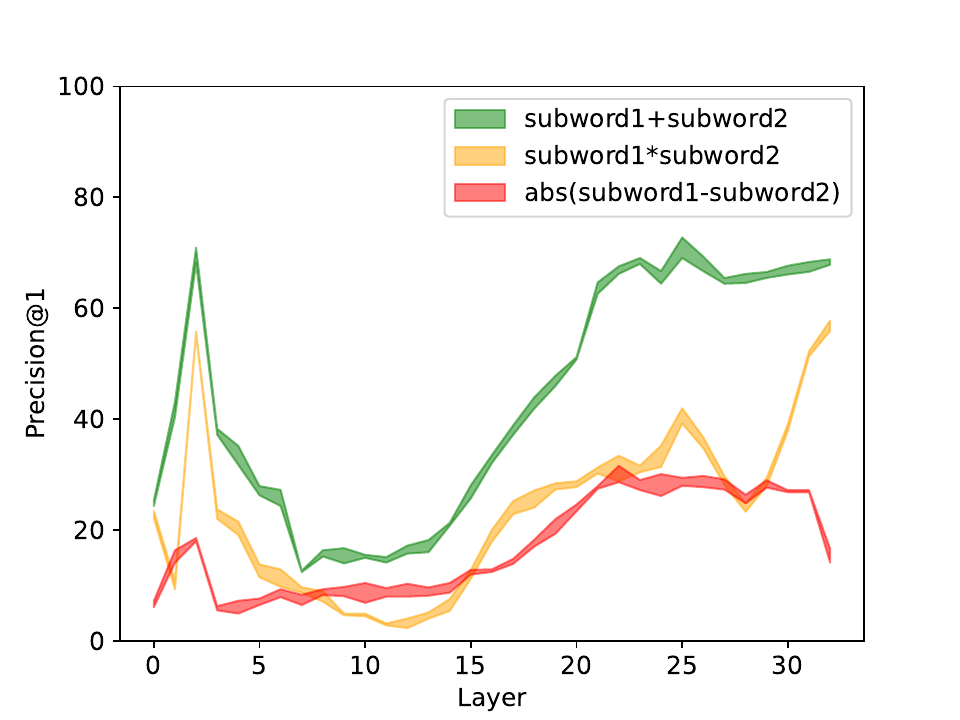}
         \caption{Aya-expanse}
    \end{subfigure}
    \hfill
    \begin{subfigure}[t]{0.49\columnwidth}
         \centering
         \includegraphics[width=\columnwidth]{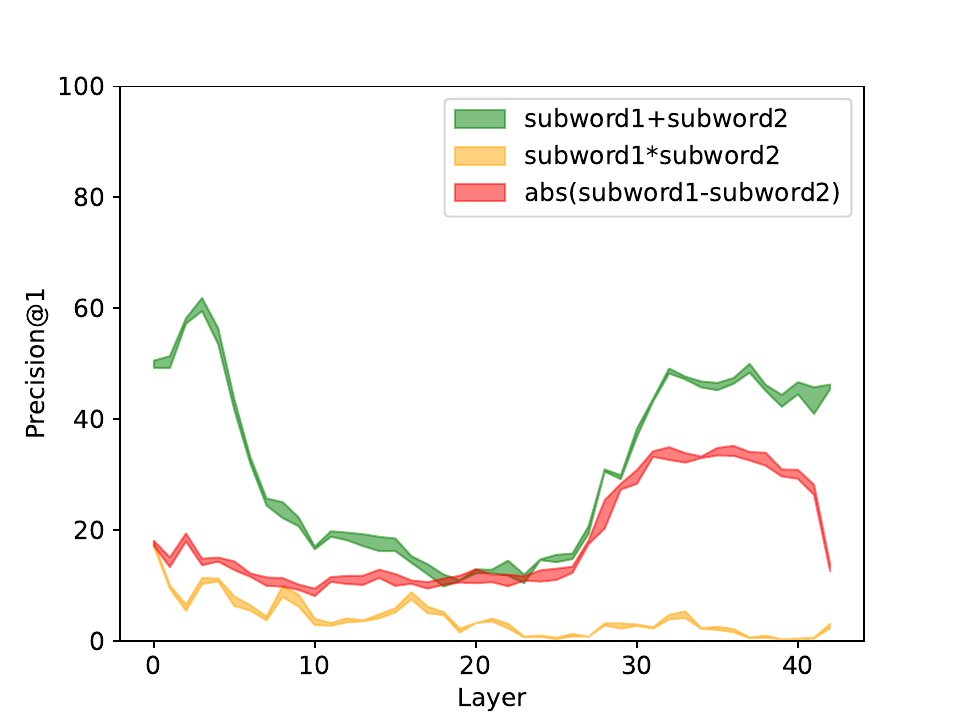}
         \caption{Gemma2}
    \end{subfigure}
    \newline
     \begin{subfigure}[t]{0.49\columnwidth}
         \centering
         \includegraphics[width=\columnwidth]{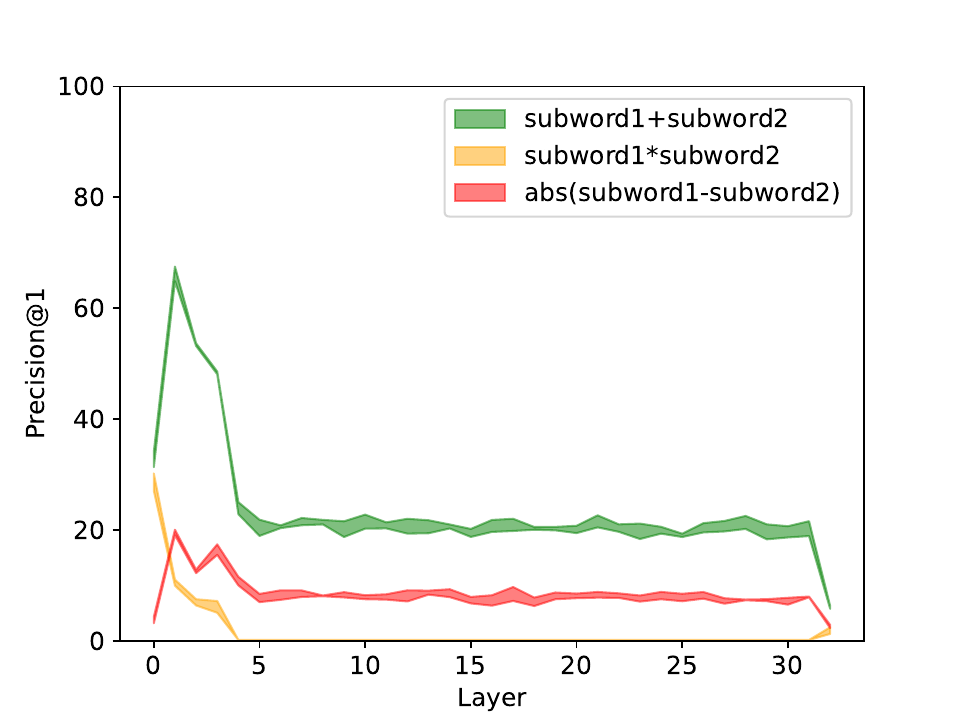}
         \caption{Falcon}
    \end{subfigure}
    \hfill
    \begin{subfigure}[t]{0.49\columnwidth}
         \centering
         \includegraphics[width=\columnwidth]{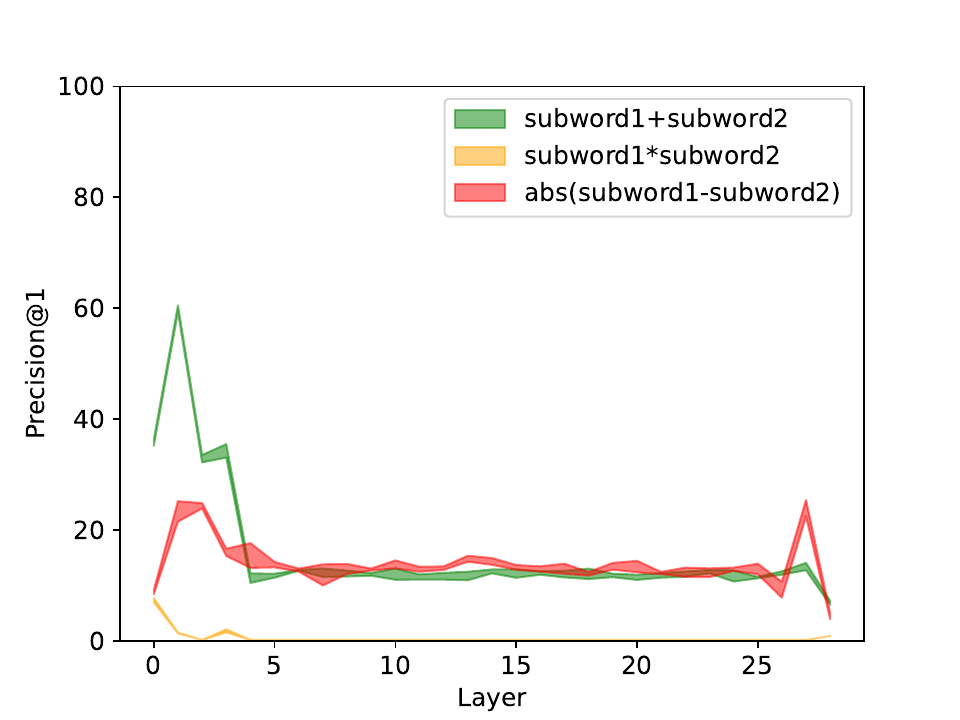}
         \caption{Qwen2.5}
    \end{subfigure}
    \vspace{-8pt}
\end{adjustbox}
\caption{Structural similarity between LLM composition and simple composition (P@1). 
Green band is simple addition. Orange refers to multiplication. Red is the performance of absolute difference. The colored bands indicate standard deviation. LLM composition is significantly more similar to simple addition.}
\label{fig:composition_operations}
\end{figure} 

A key takeaway from Figure \ref{fig:composition_operations} is that simple addition consistently outperforms other operations across all models and layers. This suggests that summing two subword representations produces a composed representation with strong structural similarity to the original whole-word representation. However, the degree of similarity varies across models, revealing three distinct patterns. Aya-expanse and Gemma2 exhibit the most impressive P@1 score, indicating high-level structural similarity between composed vectors and the original vectors. Unlike other models, the demonstrated structural similarity is able to maintain across later layers. The high precision in linear alignment exhibited in early layers of Falcon and Qwen2.5 drops in later layers. Llama models, on the other hand, only demonstrate moderate level of structural similarity between composed vectors and word vectors at the embedding layer. The structural similarity drops almost immediately.  

It is easy to see how the six LLMs can be placed in three groups with very distinct plots: Llama 3 and Llama 3.1 show very little structural similarity, and only at the embedding layer, suggesting non-linear composition or memorization. Aya-expanse and Gemma show high structural similarity, in particular at the innermost and outermost layers. Finally, Falcon and Qwen2.5 show moderate levels of structural similarity that drop last-minute. 
We discuss these differences in detail in Section \ref{sec:composition_strategy_discussion}.

\begin{figure}[!ht]
     \centering
     \begin{adjustbox}{minipage=\columnwidth,scale=1} 
     \begin{subfigure}[t]{0.49\columnwidth}
         \centering
         \includegraphics[width=\columnwidth]{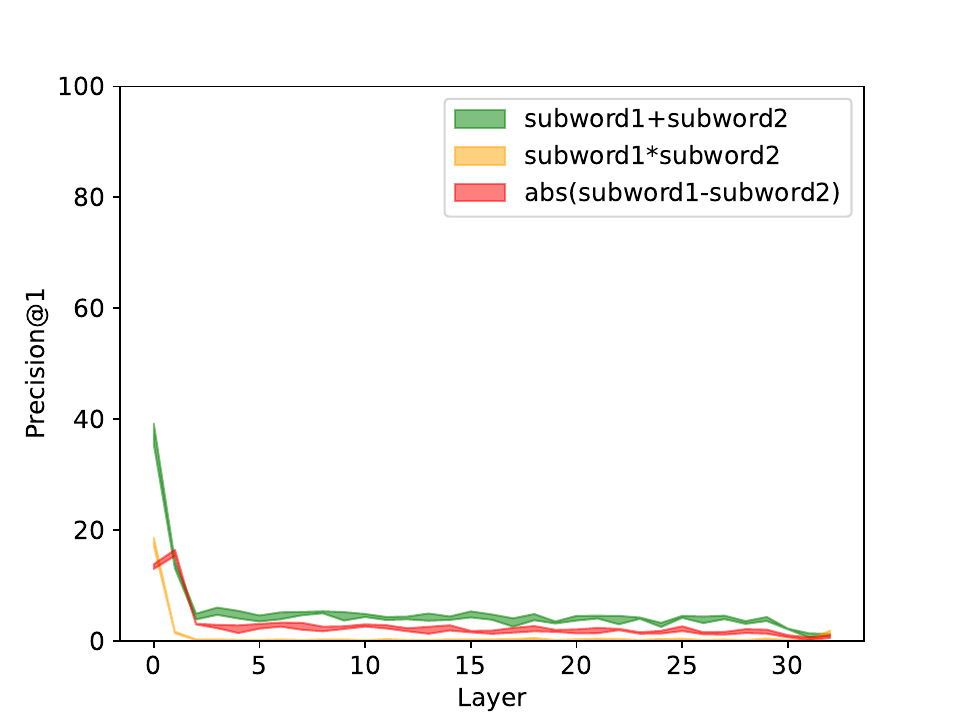}
         \caption{Llama3}
    \end{subfigure}
    \hfill
    \begin{subfigure}[t]{0.49\columnwidth}
         \centering
         \includegraphics[width=\columnwidth]{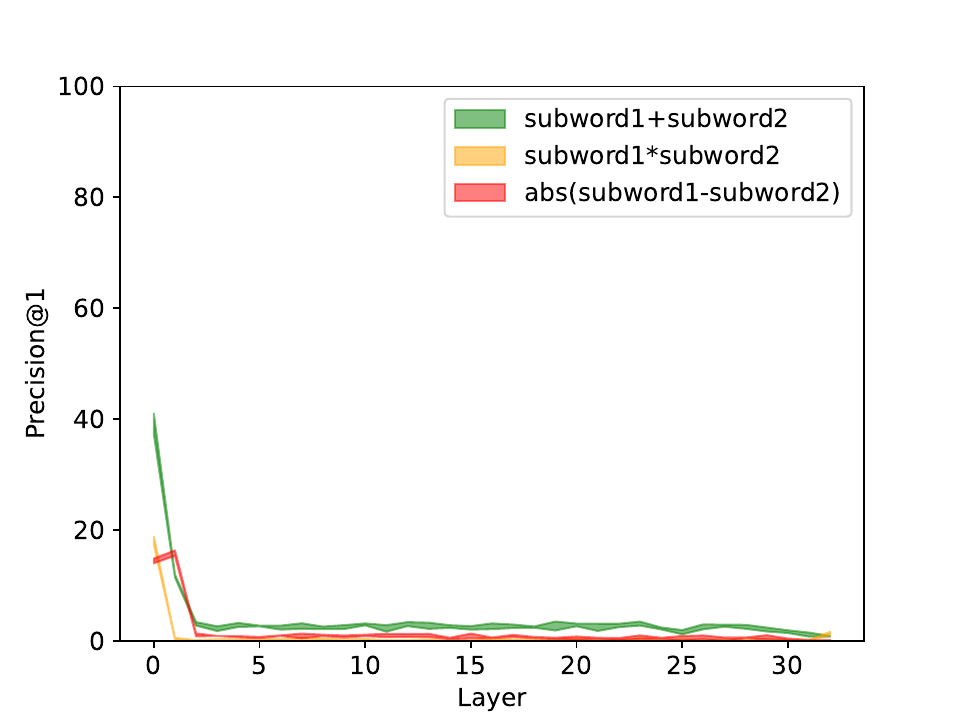}
         \caption{Llama3.1}
    \end{subfigure}
    \newline
     \begin{subfigure}[t]{0.49\columnwidth}
         \centering
         \includegraphics[width=\columnwidth]{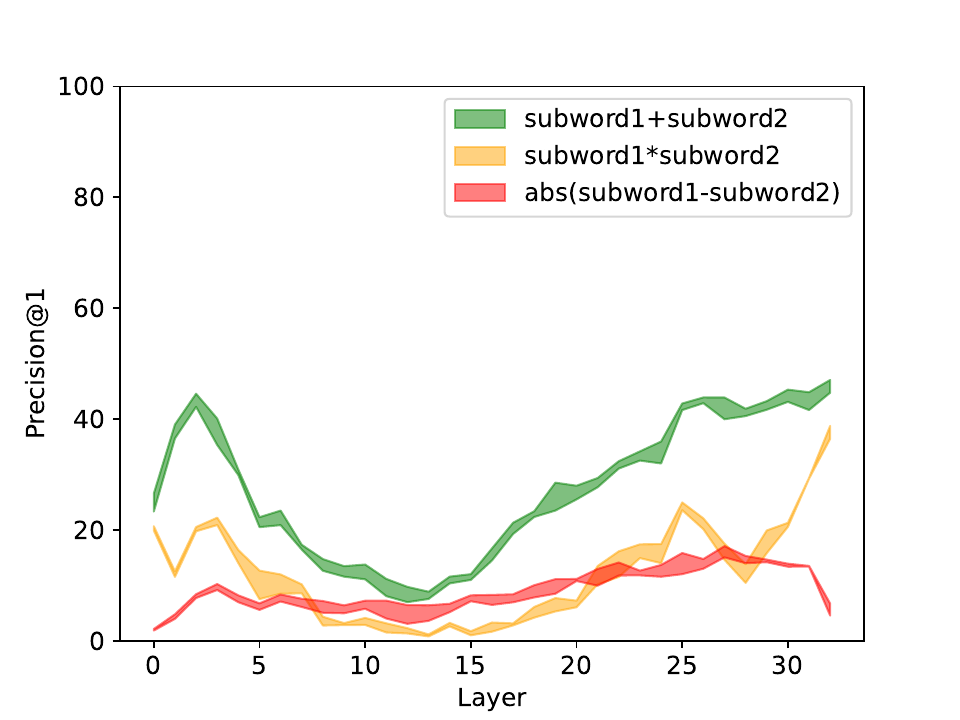}
         \caption{Aya-expanse}
    \end{subfigure}
    \hfill
    \begin{subfigure}[t]{0.49\columnwidth}
         \centering
         \includegraphics[width=\columnwidth]{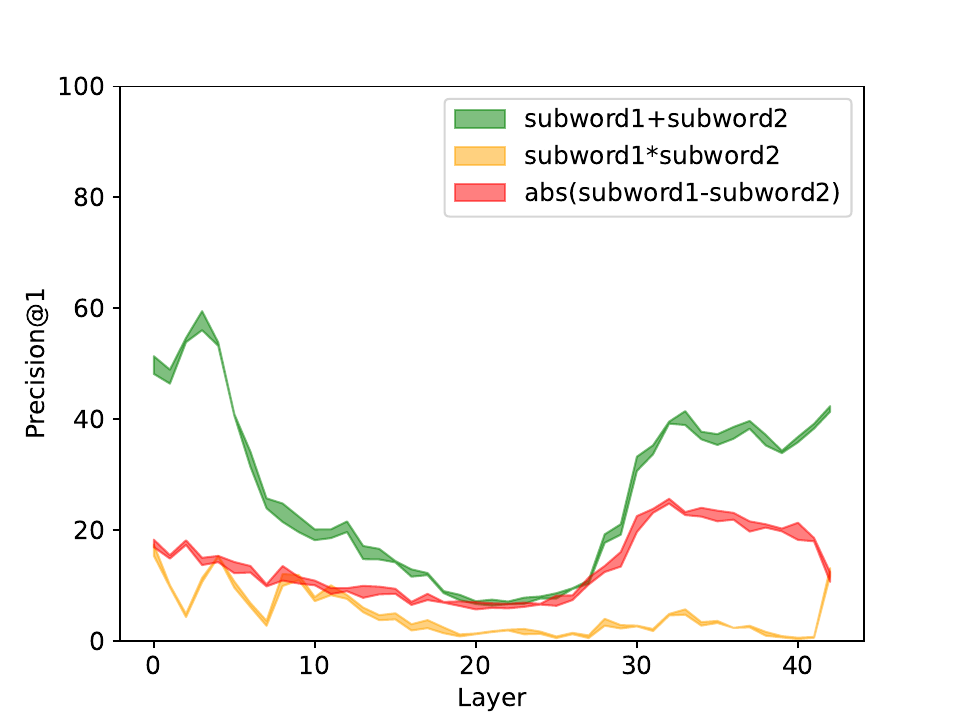}
         \caption{Gemma2}
    \end{subfigure}
    \newline
     \begin{subfigure}[t]{0.49\columnwidth}
         \centering
         \includegraphics[width=\columnwidth]{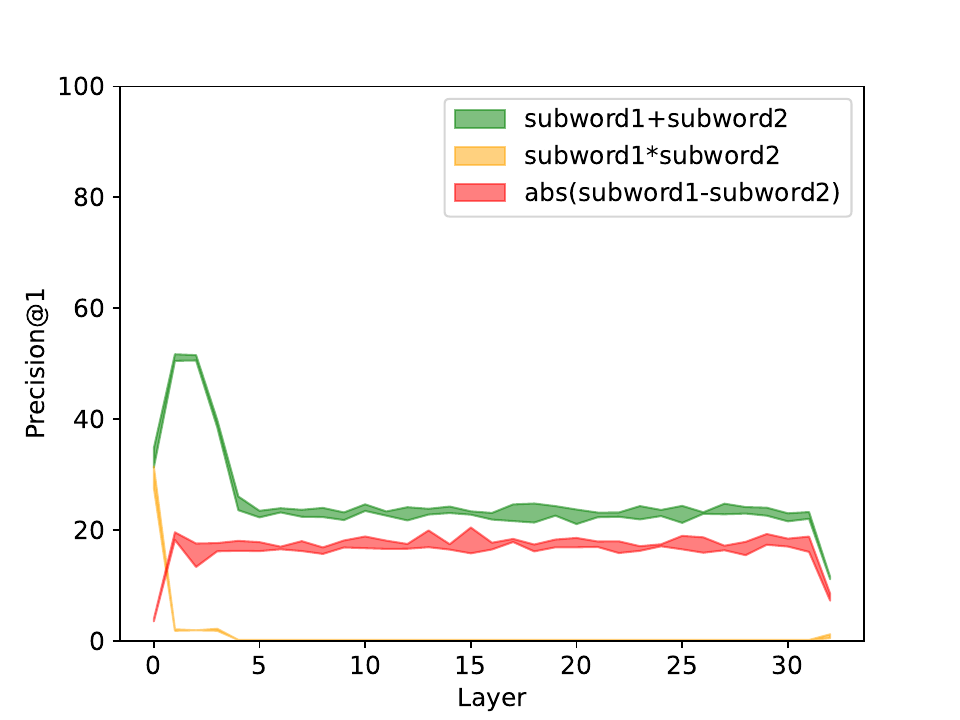}
         \caption{Falcon}
    \end{subfigure}
    \hfill
    \begin{subfigure}[t]{0.49\columnwidth}
         \centering
         \includegraphics[width=\columnwidth]{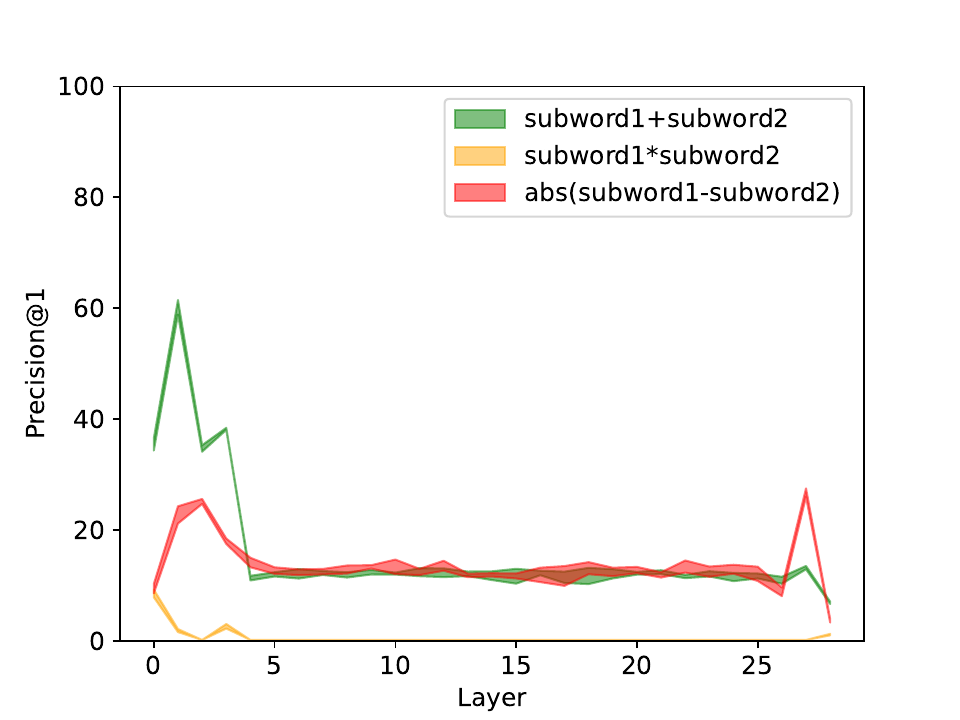}
         \caption{Qwen2.5}
    \end{subfigure}
\end{adjustbox}
\caption{Structural similarity between LLM composition (base version, not instruction-tuned) and simple composition (P@1). Green band is simple addition. Orange refers to multiplication. Red is the performance of absolute difference. The colored bands indicate standard deviation. Instruction tuning seems to have little to no impact on our results; compare with Figure~\ref{fig:composition_operations}.}
\label{fig:composition_operations_non_it}
\end{figure}

\paragraph{Impact of Instruction Tuning}
\label{sec:instruct_tuning}
All models so far were instruction-tuned. Could differences in instruction tuning explain the differences between the compositional strategies of the six LLMs? We investigate this by repeating our experiments on the base versions of the above models. This allows us to evaluate the impact on instruction-tuning on structural similarity of LLM composition and simple composition. 

The patterns in Figure \ref{fig:composition_operations_non_it} are very similar to those observed for instruction-tuned models (Figure~\ref{fig:composition_operations}). Simple addition consistently produces composed vector spaces that most closely resemble the original word vector spaces, with same three distinct groups emerging. The only small difference lies in relative performance. Structural similarity is slightly higher in instruction-tuned models compared to their base versions, while the overall patterns remain unchanged. This suggests that although instruction-tuning enhances general similarity scores, it is not the key factor driving the isometry between LLM composition and simple arithmetic operations. Instead, the structural similarity is induced during pre-training. Pre-training on large-scale corpora captures distributional and compositional regularities, inducing representations designed to facilitate composition (one way or another). What is perhaps surprising is the degree to which LLMs differ in how representations are composed. Instruction tuning improves overall similarity, but seems to merely act as a refinement process, rather than having impact on compositional strategies. 


\paragraph{Root and Non-Root Words}
The words in our dataset can be categorized into root and non-root words; see \S\ref{sec:dataset} for details. Since simple addition gave the best performance in the above, we rely on this form of composition in the following experiments. 
We now analyze how structural similarity varies across root and non-root words. Our hypothesis is that non-root words, which can be broken down into smaller meaningful units, will exhibit higher structural similarity, whereas root words, which cannot and lack obvious internal structure, will exhibit weaker alignment.

\begin{figure}[!ht]
     \centering
     \begin{adjustbox}{minipage=\columnwidth,scale=1} 
     \begin{subfigure}[t]{0.49\columnwidth}
         \centering
         \includegraphics[width=\columnwidth]{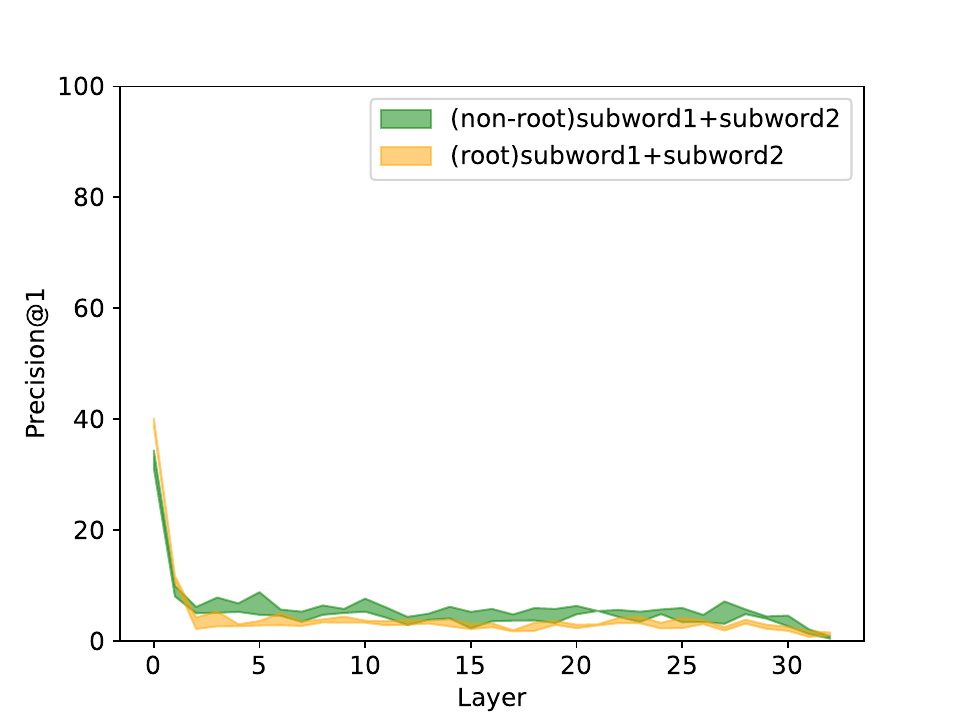}
         \caption{Llama3}
    \end{subfigure}
    \hfill
    \begin{subfigure}[t]{0.49\columnwidth}
         \centering
         \includegraphics[width=\columnwidth]{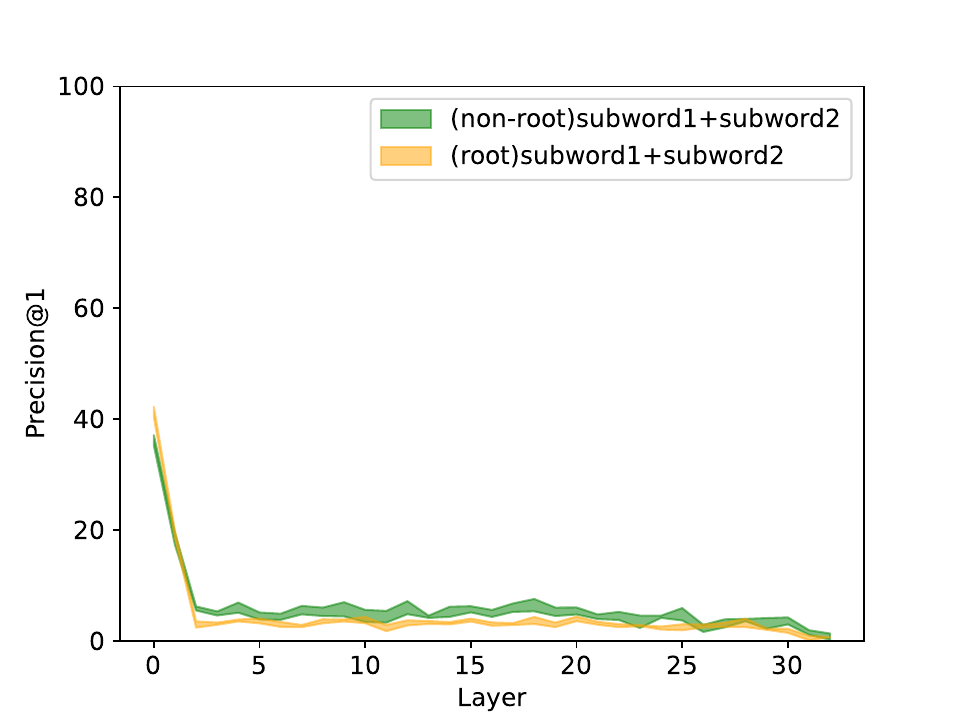}
         \caption{Llama3.1}
    \end{subfigure}
    \newline
     \begin{subfigure}[t]{0.49\columnwidth}
         \centering
         \includegraphics[width=\columnwidth]{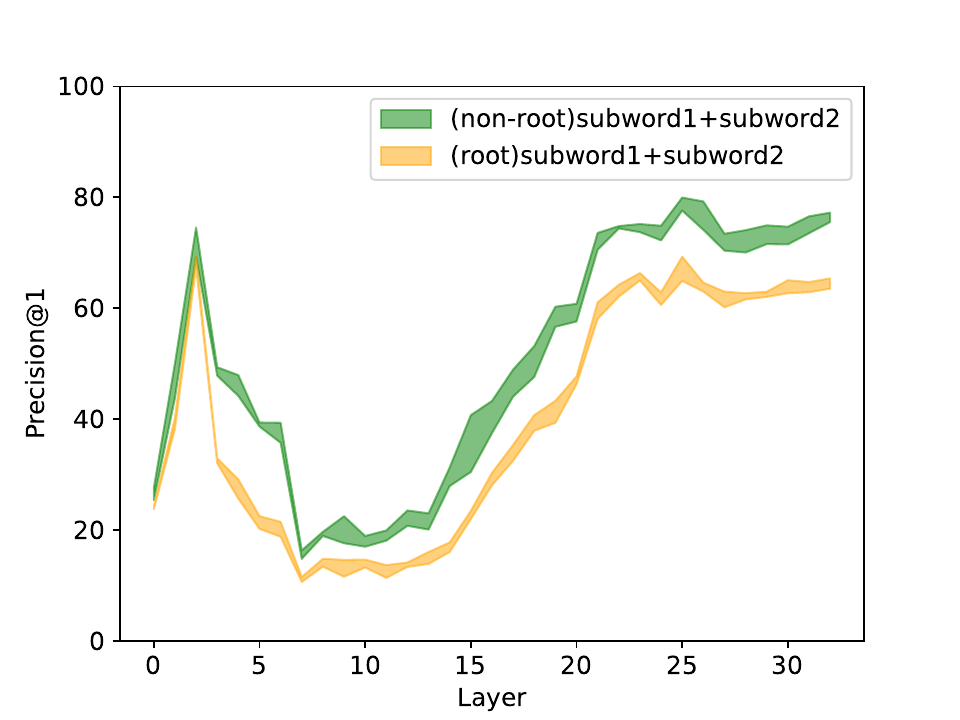}
         \caption{Aya-expanse}
    \end{subfigure}
    \hfill
    \begin{subfigure}[t]{0.49\columnwidth}
         \centering
         \includegraphics[width=\columnwidth]{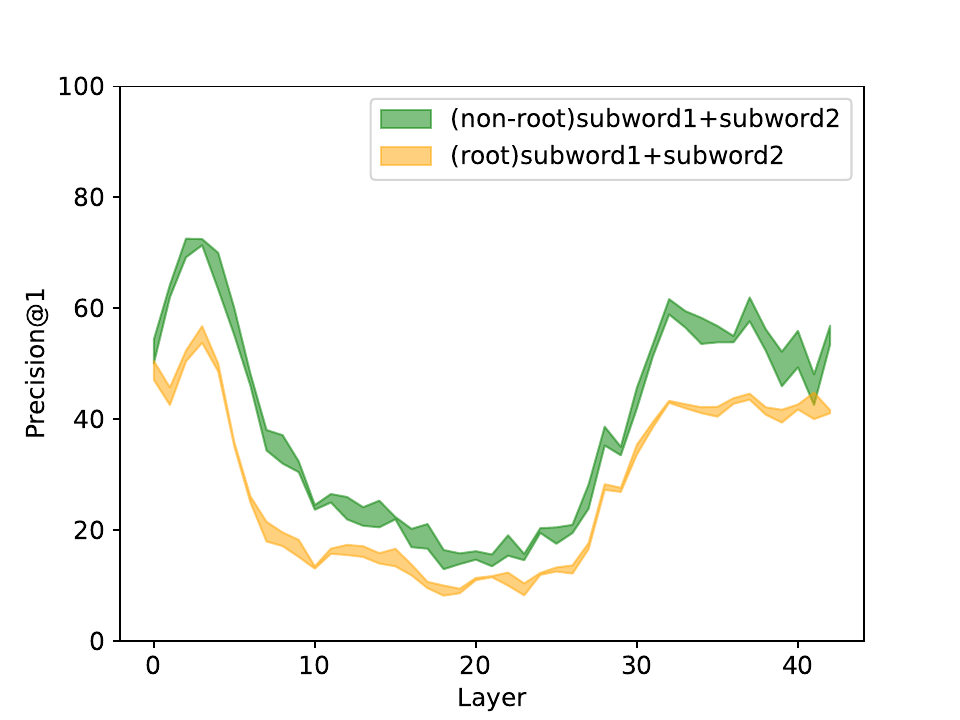}
         \caption{Gemma2}
    \end{subfigure}
    \newline
     \begin{subfigure}[t]{0.49\columnwidth}
         \centering
         \includegraphics[width=\columnwidth]{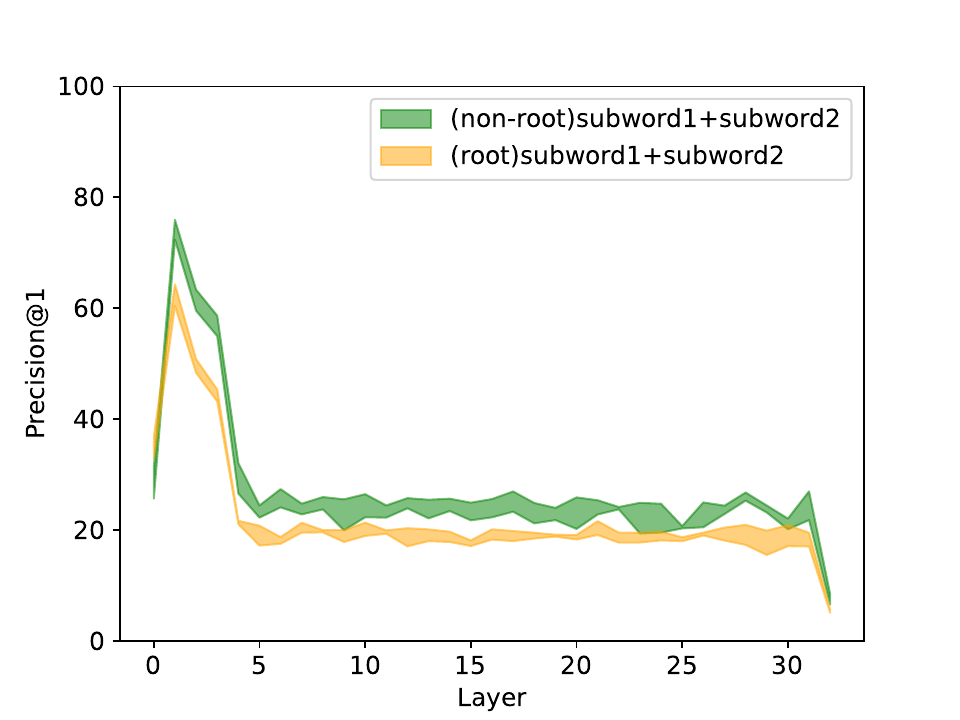}
         \caption{Falcon}
    \end{subfigure}
    \hfill
    \begin{subfigure}[t]{0.49\columnwidth}
         \centering
         \includegraphics[width=\columnwidth]{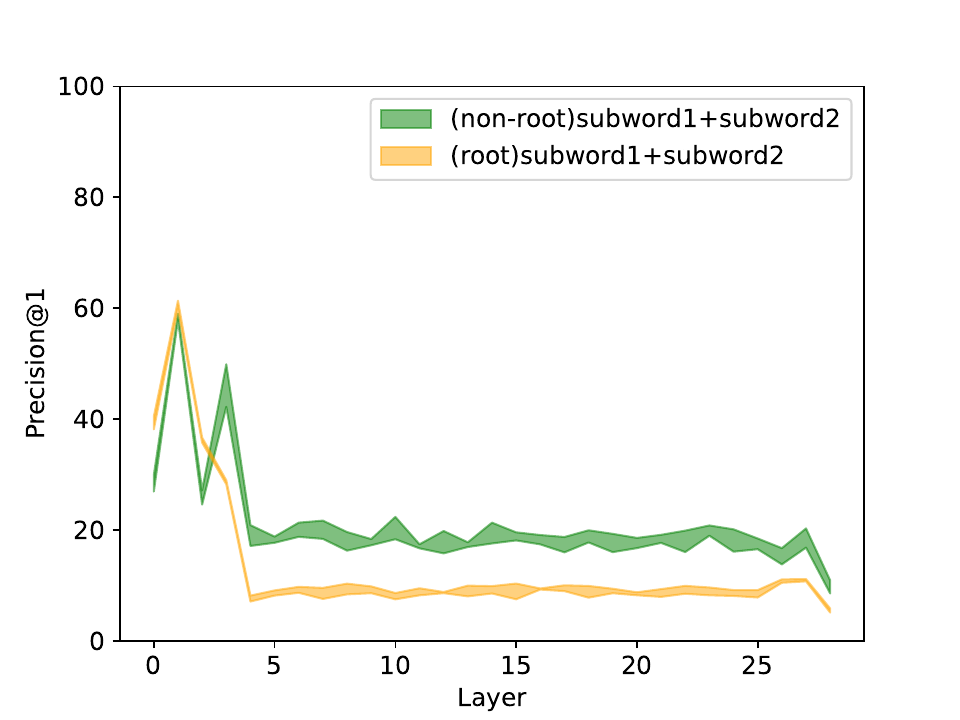}
         \caption{Qwen2.5}
    \end{subfigure}
\end{adjustbox}
\caption{Structural similarity between LLM composition and simple composition (P@1) across all layers and different word types. Green refers to the performance on non-root words. Orange refers to root words.}
\label{fig:geometry_word_type}
\vspace{-10pt}
\end{figure} 

Figure \ref{fig:geometry_word_type} illustrates that across different models and layers, non-root words consistently exhibit higher structural similarity than root words. This suggests that simple addition more effectively produces a composed vector representation that aligns linearly with the original word representation for non-root words. This was expected and lends support to our original hypothesis. In contrast, root words exhibit weaker linear alignment, likely because they function as semantic atoms that are not easily decomposed into smaller parts in a meaningful way. Since their meanings are not derived from the interaction of multiple components, their representations may be shaped more by contextual factors and usage patterns than by explicit compositional relationships. This could introduce greater variability in their spatial organization, leading to generally lower structural similarity.

\begin{figure}[!ht]
     \centering
     \begin{adjustbox}{minipage=\columnwidth,scale=1} 
     \begin{subfigure}[t]{0.49\columnwidth}
         \centering
         \includegraphics[width=\columnwidth]{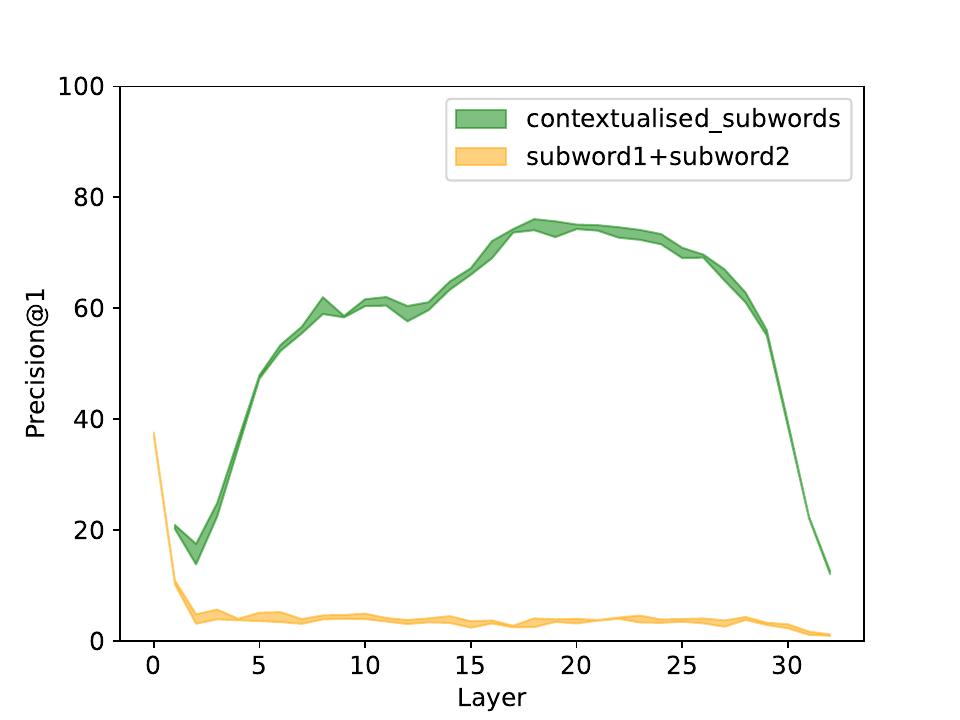}
         \caption{Llama3}
    \end{subfigure}
    \hfill
    \begin{subfigure}[t]{0.49\columnwidth}
         \centering
         \includegraphics[width=\columnwidth]{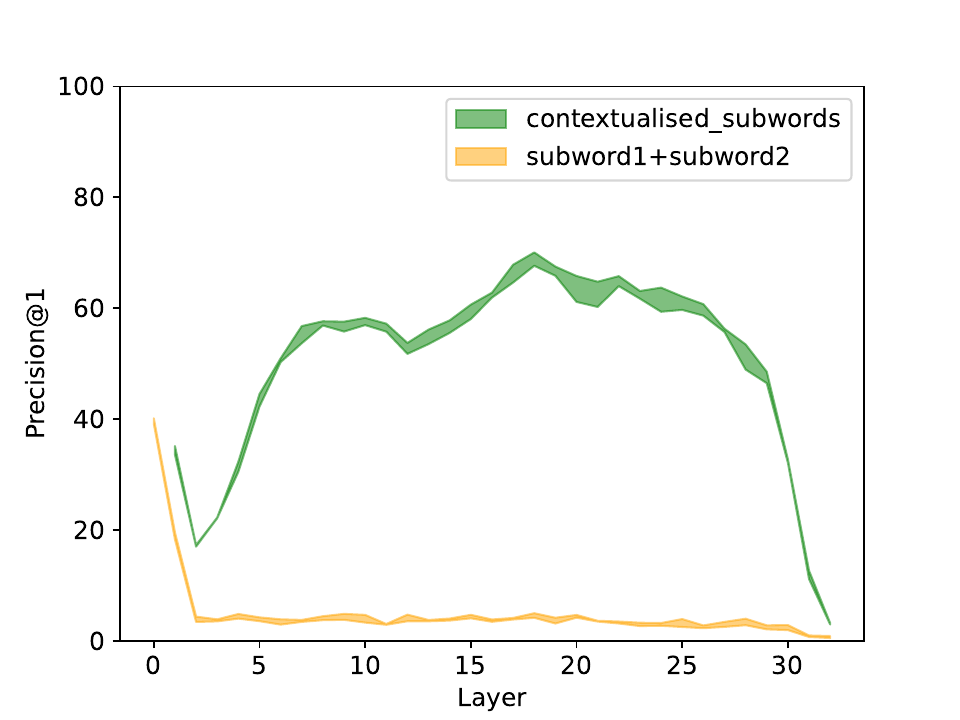}
         \caption{Llama3.1}
    \end{subfigure}
    \newline
     \begin{subfigure}[t]{0.49\columnwidth}
         \centering
         \includegraphics[width=\columnwidth]{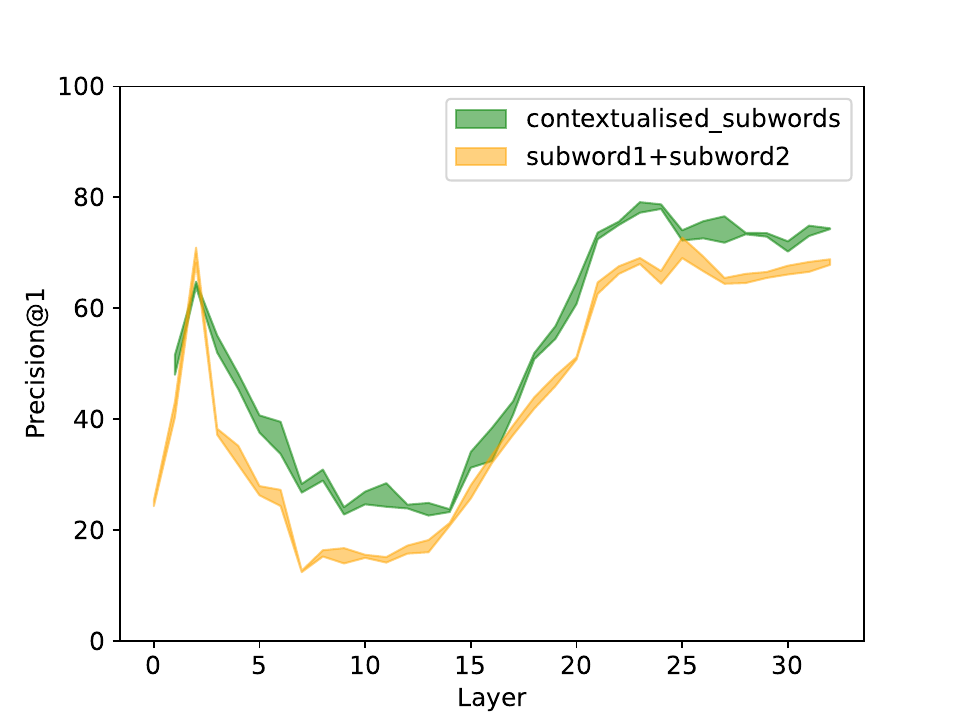}
         \caption{Aya-expanse}
    \end{subfigure}
    \hfill
    \begin{subfigure}[t]{0.49\columnwidth}
         \centering
         \includegraphics[width=\columnwidth]{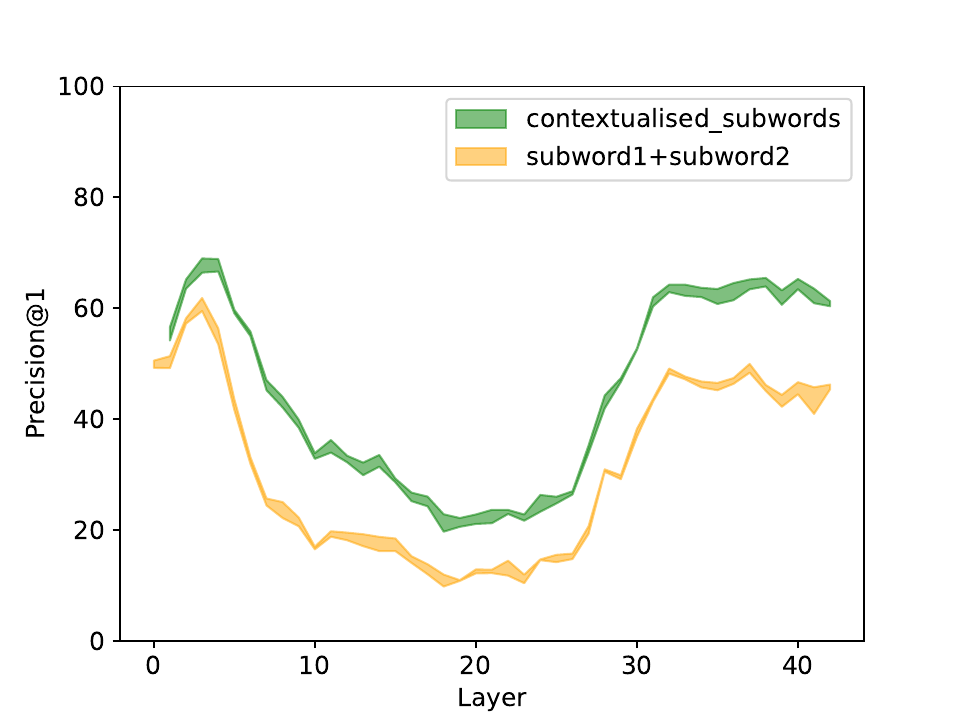}
         \caption{Gemma2}
    \end{subfigure}
    \newline
     \begin{subfigure}[t]{0.49\columnwidth}
         \centering
         \includegraphics[width=\columnwidth]{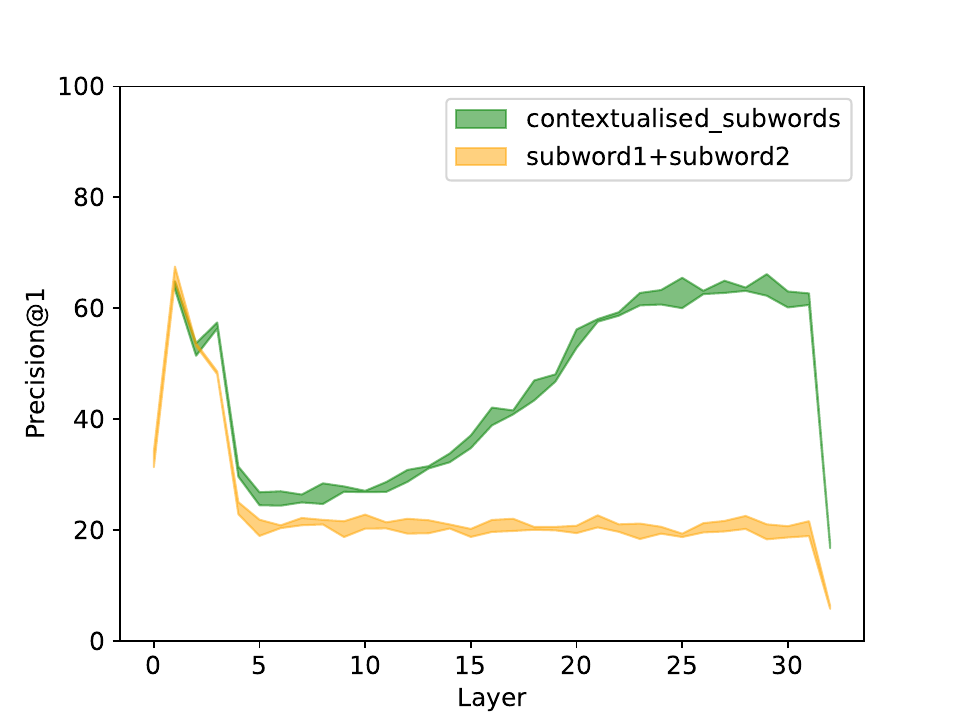}
         \caption{Falcon}
    \end{subfigure}
    \hfill
    \begin{subfigure}[t]{0.49\columnwidth}
         \centering
         \includegraphics[width=\columnwidth]{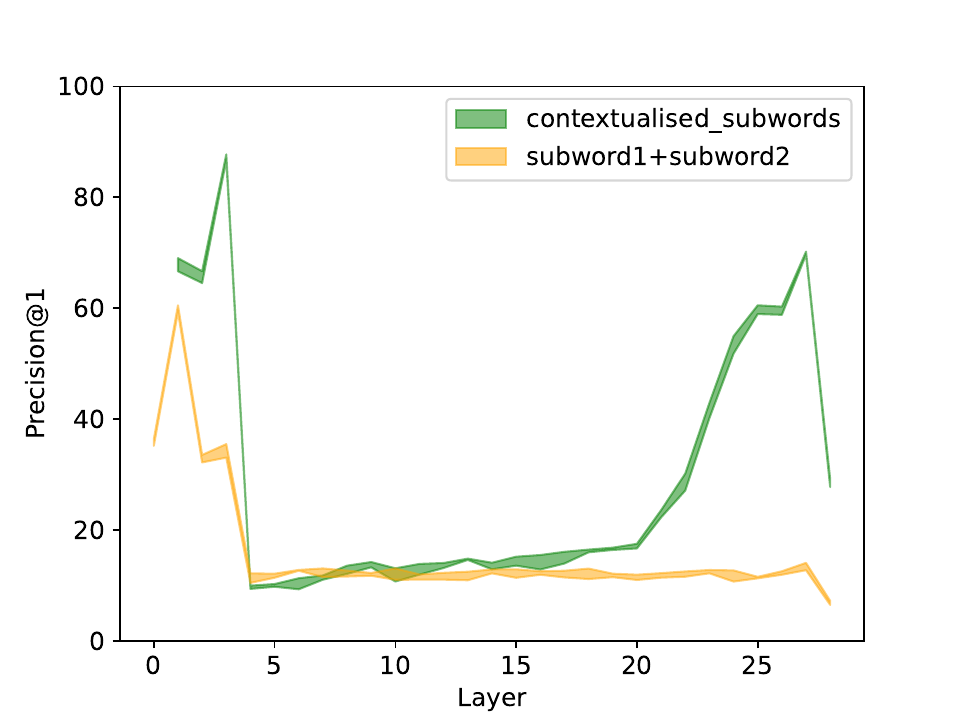}
         \caption{Qwen2.5}
    \end{subfigure}
\end{adjustbox}
\caption{Structural similarity between LLM composition and simple composition (P@1) across all layers w/wo contextualization. Green refers to the performance with contextualization. Orange refers to without contextualization.}
\label{fig:geometry_contextualization}
\vspace{-10pt}
\end{figure}

\begin{figure*}[!ht]
\vspace{-10pt}
     \centering
     \begin{adjustbox}{minipage=\textwidth,scale=0.8} 
    \begin{subfigure}[t]{0.32\textwidth}
         \centering
         \includegraphics[width=\textwidth]{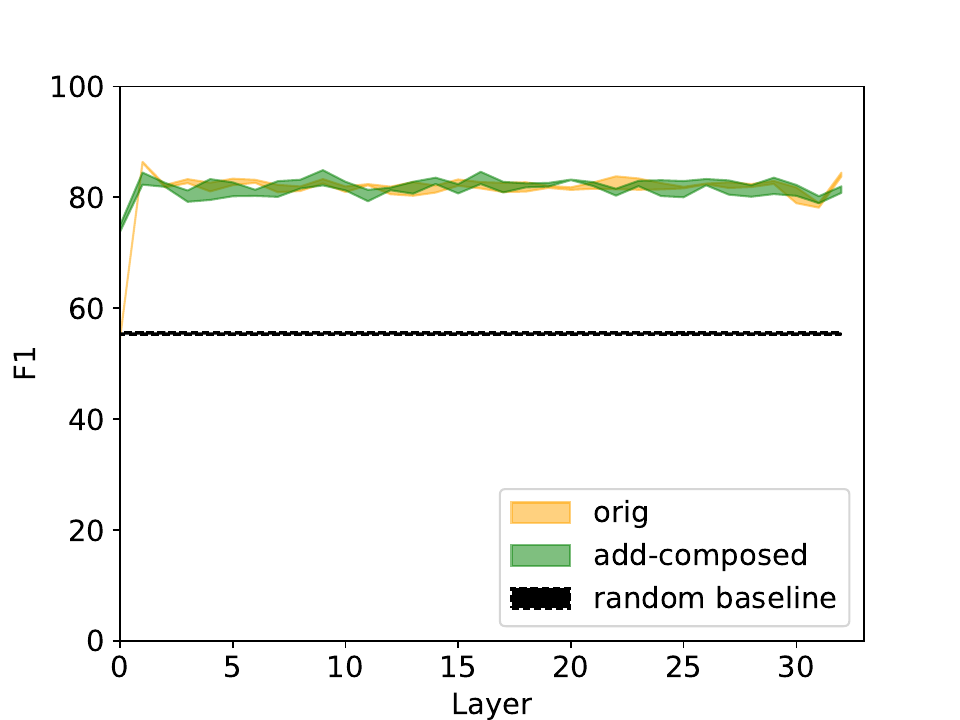}
         \caption{Llama3}
     \end{subfigure}
     \hfill
     \begin{subfigure}[t]{0.32\textwidth}
         \centering
         \includegraphics[width=\textwidth]{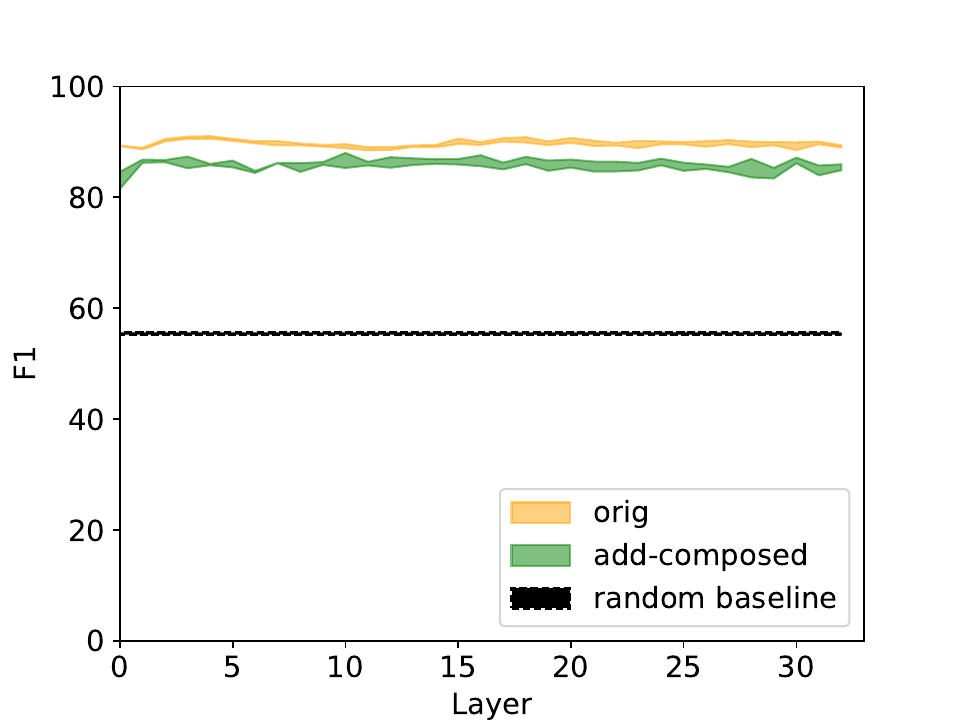}
         \caption{Aya-expanse} 
     \end{subfigure}
    \hfill
    \begin{subfigure}[t]{0.32\textwidth}
         \centering
         \includegraphics[width=\textwidth]{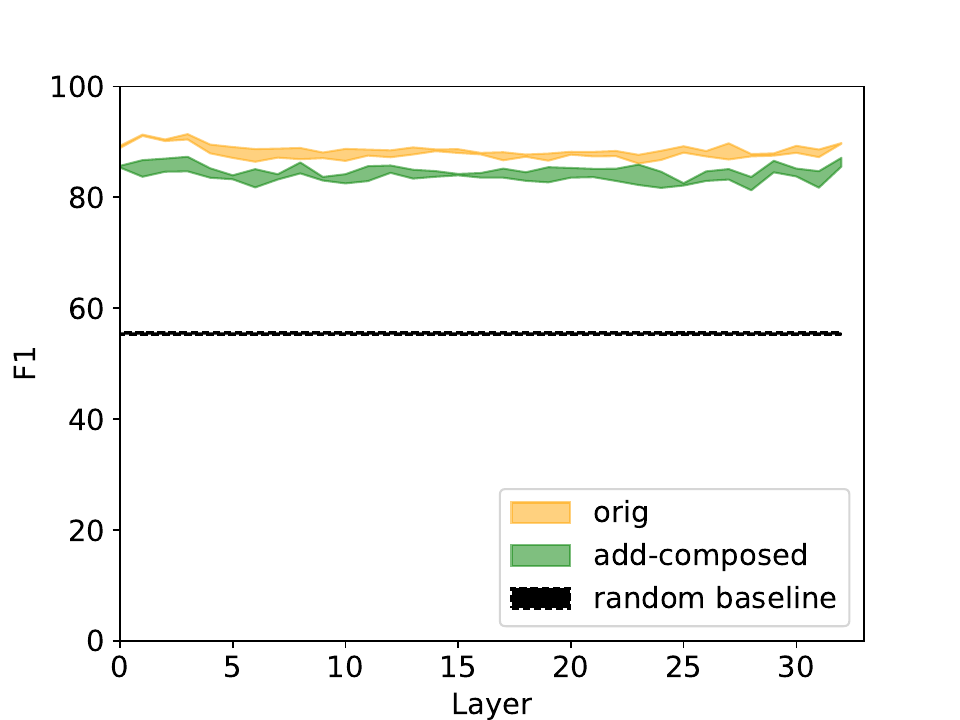}
         \caption{Falcon}
    \end{subfigure}
    \newline
     \begin{subfigure}[t]{0.32\textwidth}
         \centering
         \includegraphics[width=\textwidth]{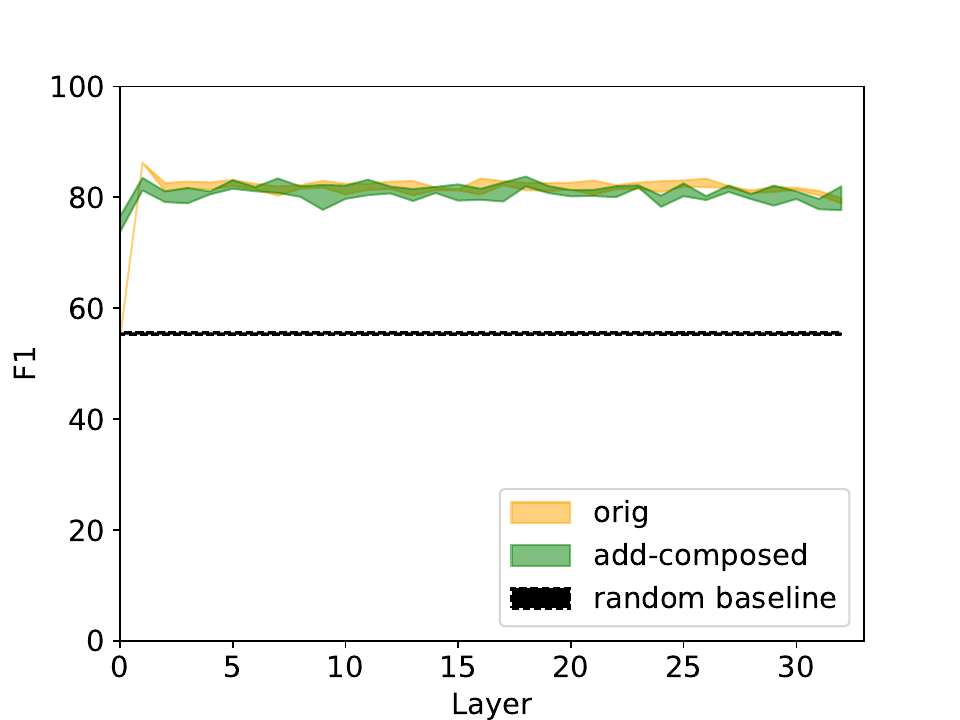}
         \caption{Llama3.1}
     \end{subfigure}
     \hfill
     \begin{subfigure}[t]{0.32\textwidth}
         \centering
         \includegraphics[width=\textwidth]{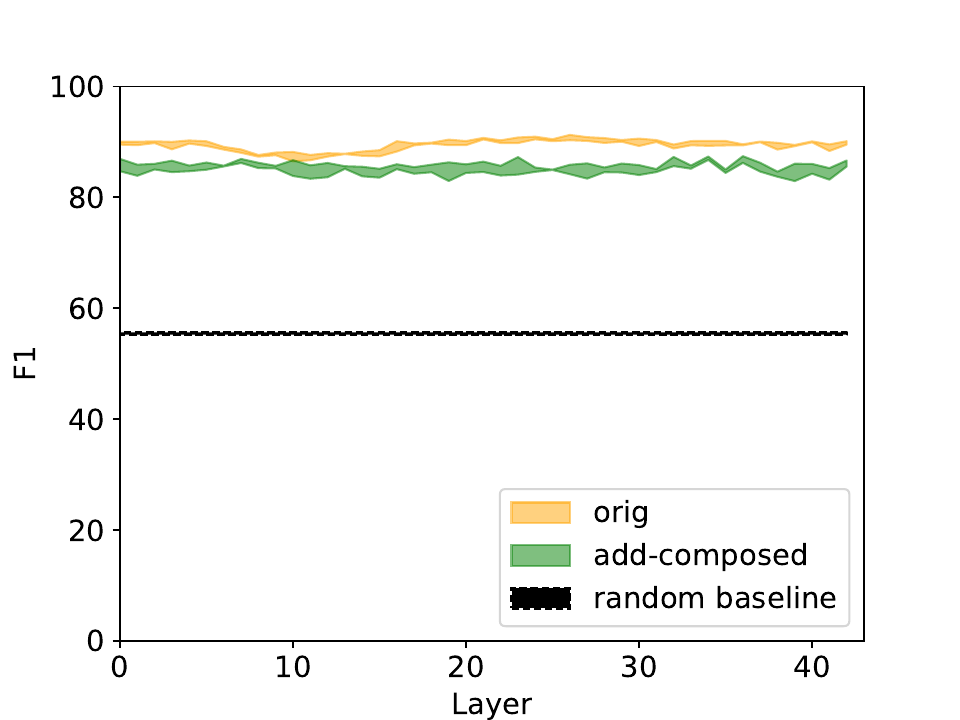}
         \caption{Gemma2} 
     \end{subfigure}
    \hfill
    \begin{subfigure}[t]{0.32\textwidth}
         \centering
         \includegraphics[width=\textwidth]{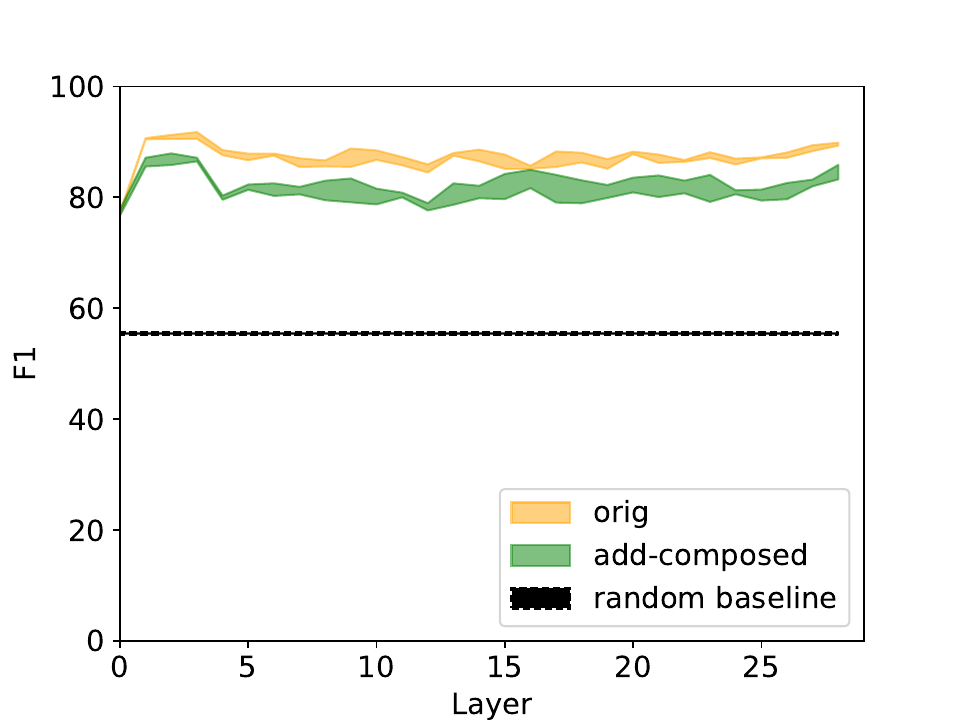}
         \caption{Qwen2.5}
    \end{subfigure}
\end{adjustbox}
\caption{Performance (weighted F1) of different LLMs on word type classification across all layers. Orange indicates the performance of using the original whole words. Green refers to addition-composed performance, and black is the random baseline. The colored bands indicate standard deviation.}         
\label{fig:word_content_prediction}
\vspace{-5pt}
\end{figure*} 

\paragraph{Impact of Contextualization}
Previous experiments have investigated the structural similarity between composed vectors—formed by combining two separate static subword representations—and original word representations. Many recent approaches using LLMs produce word, phrase, or sentence embeddings by applying mean pooling over their contextualized token representations. In this experiment, we take a similar approach by feeding both subwords into the LLM simultaneously, allowing their representations to interact and refine with each other. We then examine whether a simple addition of these contextualized subword representations can effectively reconstruct a composed representation that maintains structural similarity to the original word representation.



Figure \ref{fig:geometry_contextualization} compares results with and without contextualization. When contextualization is applied, all models exhibit stronger linear alignment across layers. Notably, Llama models, along with Falcon and Qwen2.5, display distinct patterns. Instead of showing minimal structural similarity, Llama models demonstrate high levels of isometry in their middle layers. Falcon and Qwen2.5 also achieve higher P@1 scores in later layers. Meanwhile, Aya and Gemma models maintain a pattern similar to the non-contextualized scenario, but with generally higher structural similarity. These findings suggest that for some LLMs, e.g., Llama and Llama3.1, composed representations are only similar to simply arithmetic compositions 
when the LLM has observed both subwords in the same context. This highlights two distinct composition mechanisms. The first, seen in Aya and Gemma2, allows a linearly alignable composed representation to be directly formed by adding the separate subword representations. The second, observed in Llama, requires the model to process the subwords in the same context before producing a linearly alignable composed representation, possibly indicating higher degrees of memorization.  

\section{Probing Analysis}
Previous geometry experiments have demonstrated that there exists a high degree of structural similarity between composed representations and the whole-word representations. However, this structural similarity varies across layers and models. In the following experiments, we investigate whether some basic aspects of the word understanding, specially content and form, have been preserved in the composed representation. 

\subsection{Root and Non-Root Words}
As shown in Table \ref{sec:dataset}, words in the dataset can be classified into root and non-root words. Identifying whether a given vector representation corresponds to a root or non-root word requires capturing content information. Root words are the smallest meaningful units that cannot be broken down further, whereas non-root words are decomposable in meaning.

\paragraph{Method} This word type prediction task is framed as a binary classification problem. We train a simple logistic regression model using either the original word representations or the composed subword representations as input. The classifier is trained for three epochs with a batch size of 8, utilizing the Adam optimizer with a learning rate of 1e-3.  

\paragraph{Results} The experiment results, measured by the weighted F1 score, are summarized in Figure \ref{fig:word_content_prediction}. The orange line represents the weighted F1 score across all layers using the original word representations as input, while the green line shows the performance when using composed representations obtained by summing two subword representations.

Preliminary experiments indicate that a random baseline (black line) would achieve approximately 56\% weighted F1 score. In contrast, features extracted from composed representations enable the model to achieve over 80\% weighted F1 score, demonstrating that the distinction between root and non-root words is inherently embedded in the composed representations. The small gap between the green and original lines further suggests a high degree of content information preservation. 

Despite the variations in structural similarity observed in previous geometry analysis, composed representations maintain consistently high performance across different layers in the word type prediction task. This implies that even if a composed representation does not perfectly align with the original word representation in vector space, it could still preserve essential semantic information about the word.

\begin{figure*}[!ht]
     \centering
     \begin{adjustbox}{minipage=\textwidth,scale=0.8} 
    \begin{subfigure}[t]{0.32\textwidth}
         \centering
         \includegraphics[width=\textwidth]{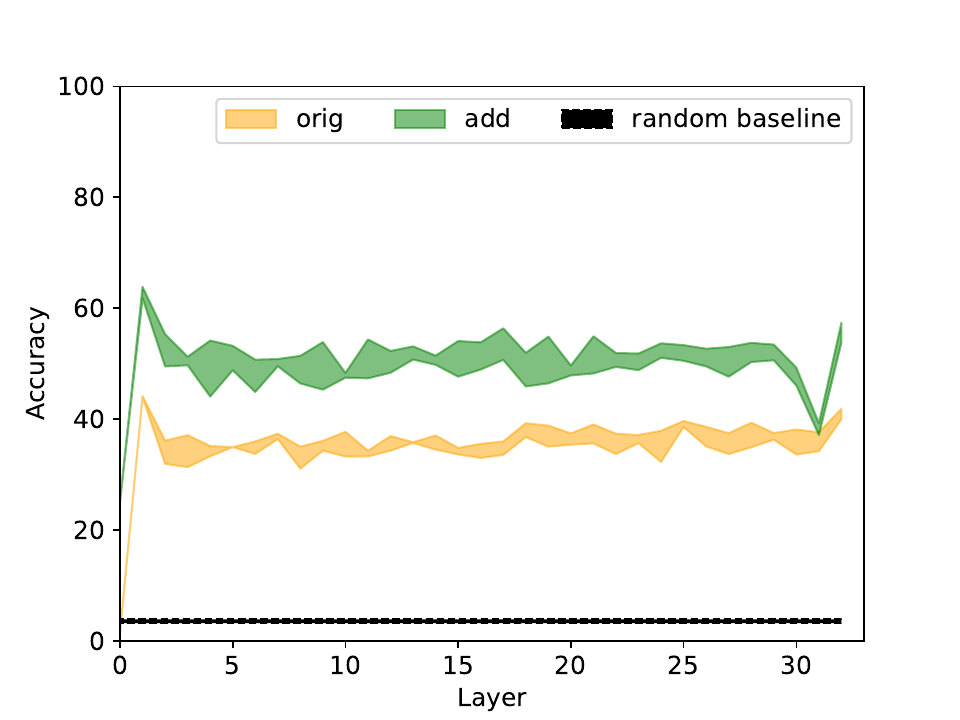}
         \caption{Llama3}
     \end{subfigure}
     \hfill
     \begin{subfigure}[t]{0.32\textwidth}
         \centering
         \includegraphics[width=\textwidth]{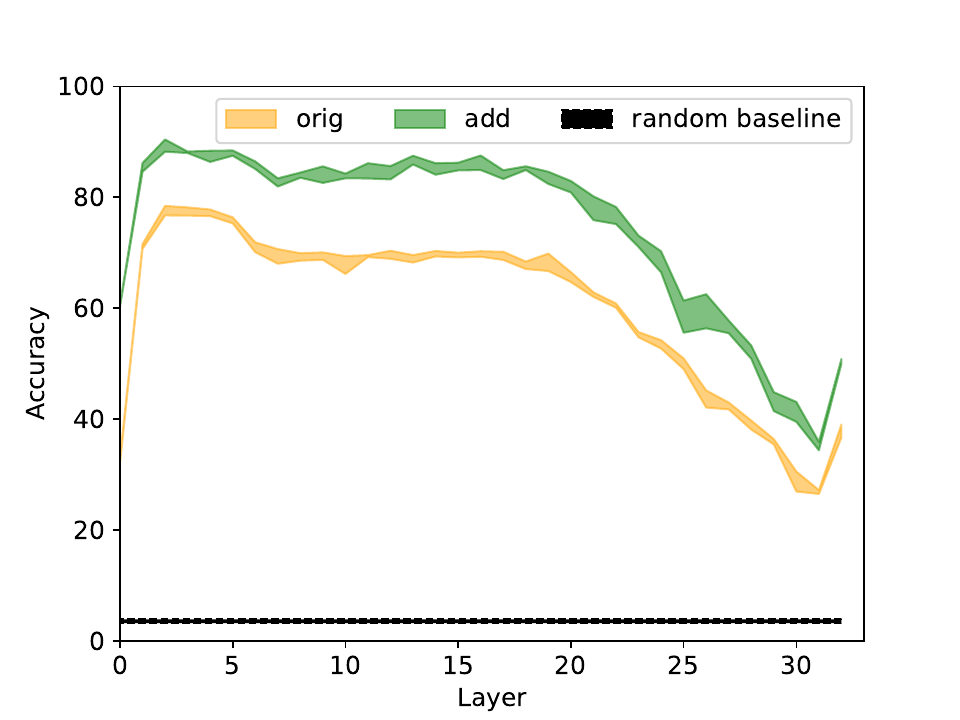}
         \caption{Aya-expanse} 
     \end{subfigure}
    \hfill
    \begin{subfigure}[t]{0.32\textwidth}
         \centering
         \includegraphics[width=\textwidth]{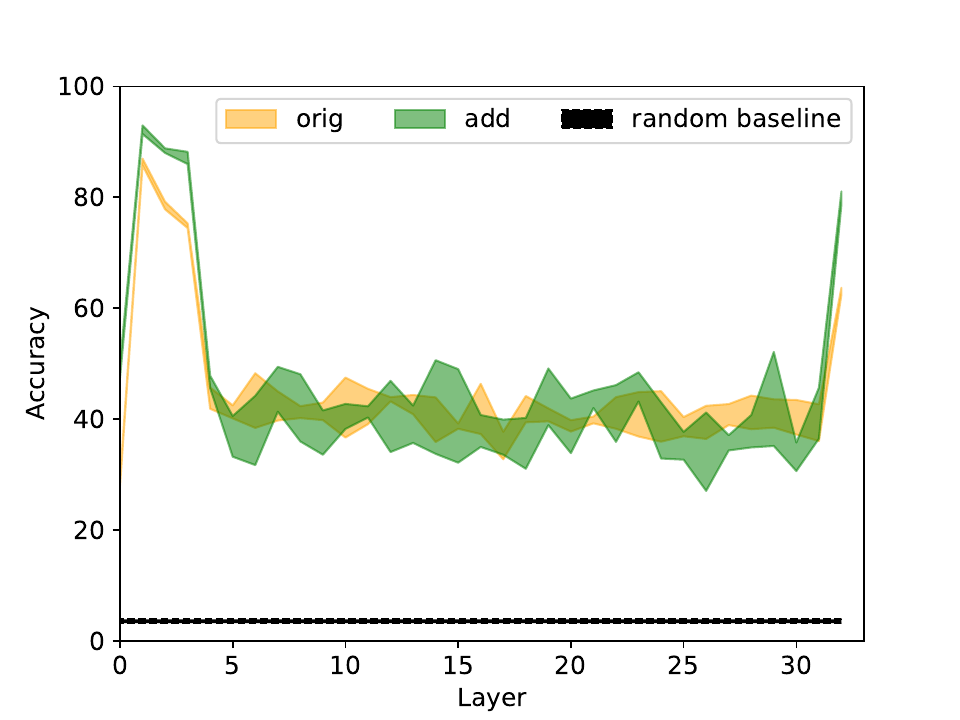}
         \caption{Falcon}
    \end{subfigure}
    \newline
     \begin{subfigure}[t]{0.32\textwidth}
         \centering
         \includegraphics[width=\textwidth]{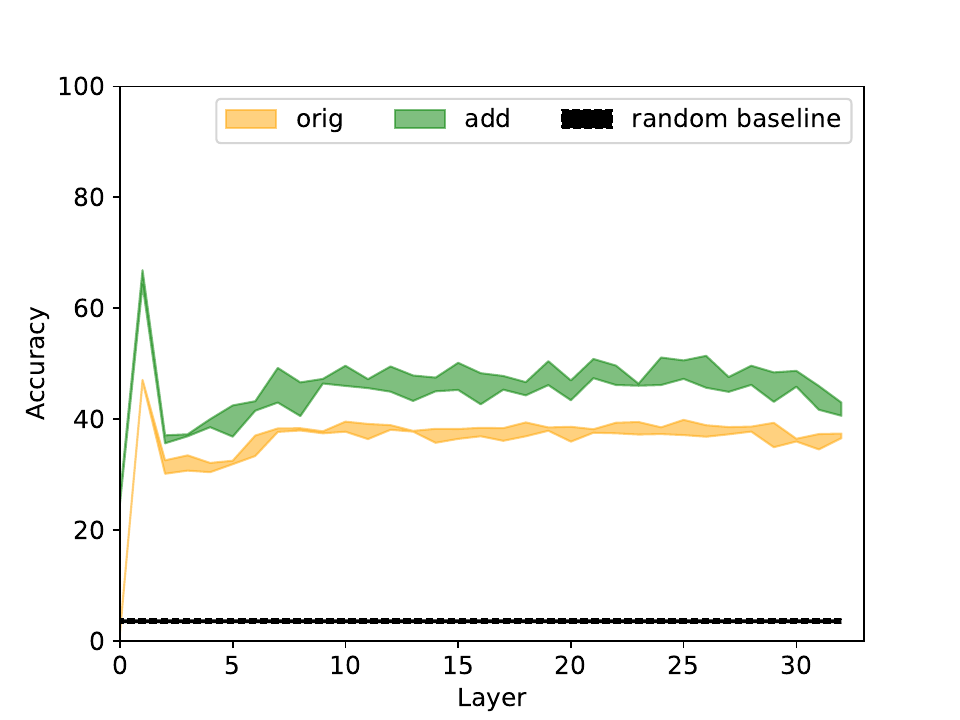}
         \caption{Llama3.1}
     \end{subfigure}
     \hfill
     \begin{subfigure}[t]{0.32\textwidth}
         \centering
         \includegraphics[width=\textwidth]{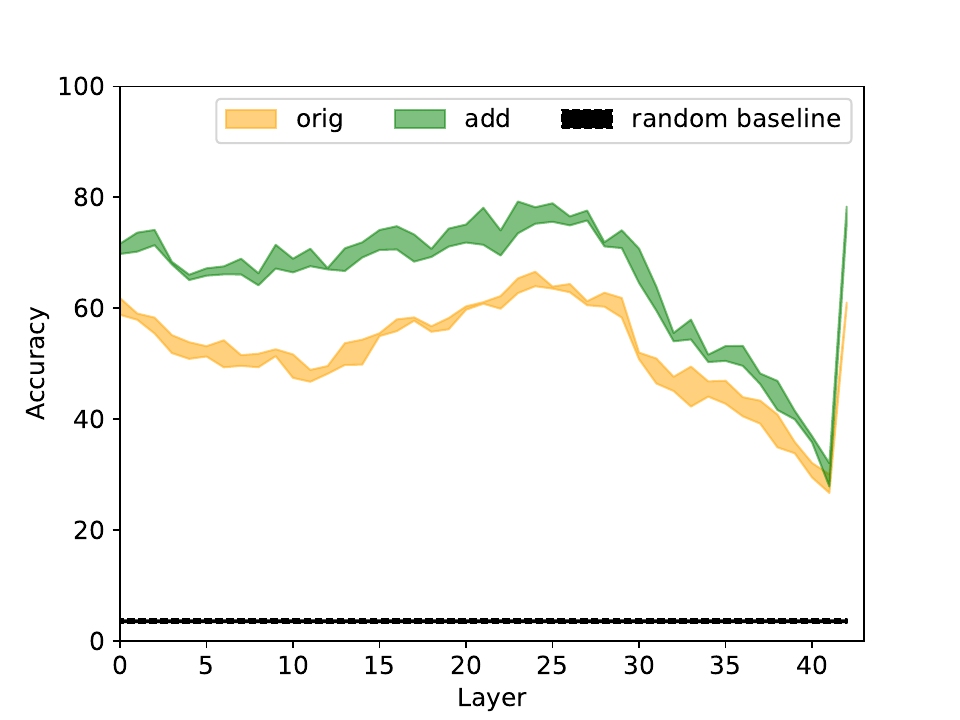}
         \caption{Gemma2} 
     \end{subfigure}
    \hfill
    \begin{subfigure}[t]{0.32\textwidth}
         \centering
         \includegraphics[width=\textwidth]{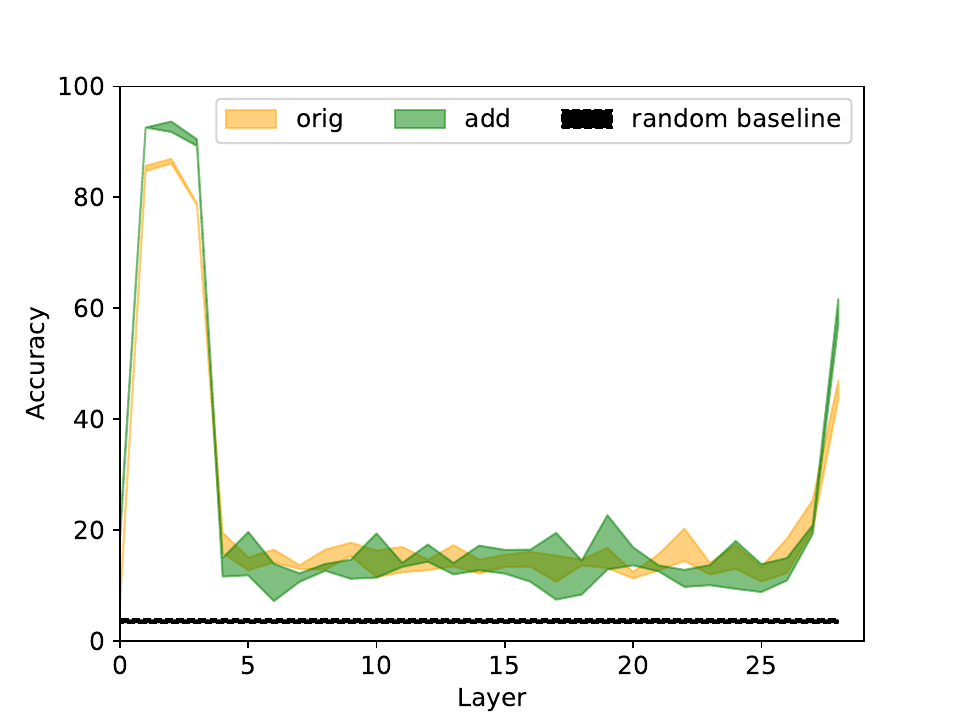}
         \caption{Qwen2.5}
    \end{subfigure}
\end{adjustbox}
\caption{Performance (Accuracy) of different LLMs on word length prediction across all layers. Orange indicates the performance of using the original whole words. Green refers to addition-composed performance, and black is the random baseline. The colored bands indicate standard deviation.}         
\label{fig:word_form_prediction}
\vspace{-5pt}
\end{figure*} 

\subsection{Word Length Prediction}
Having considered semantic classes, we now investigate whether LLMs retain form-related properties of subword constituents, specifically whether information about {\em word length} is passed up the network. Similar to our earlier experiment, we assess whether this information is encoded by predicting word length from both original and composed representations.

\paragraph{Method} We formulate word length prediction as a regression task. Using linear regression, we predict the word length from a given vector representation. The regressor is trained for three epochs with a batch size of 8, using Adam optimizer with a learning rate of 1e-3. Since word length is a discrete value, the predicted outputs are rounded to the nearest integer before computing accuracy.

\paragraph{Results} Figure \ref{fig:word_form_prediction} presents the overall accuracy across different models and layers. A random baseline (black line) achieves approximately 3.5\% accuracy, reflecting the difficulty of the task without meaningful features. In contrast, both original word representations and composed subword representations result in significantly higher accuracy, demonstrating that word length information is inherently encoded in these embeddings.

Across all six LLMs, a consistent pattern emerges: the highest accuracy is observed in the early layers, suggesting that form-related properties, such as word length, are well-preserved at lower levels of the representation. However, as layers deepen, accuracy gradually decreases, likely due to the increasing abstraction of form information. Interestingly, at the final layers, accuracy improves again, indicating that some form-related information re-emerges at later processing stages. This suggests that while middle layers prioritize semantic abstraction, early and late layers retain more explicit surface-level features.

Consistent with the geometry analysis, these six models can be grouped into the same three categories based on their layer-wise accuracy patterns: (1) Llama3 and Llama3.1, (2) Aya and Gemma2, and (3) Falcon and Qwen2.5. The similar trends observed across models reinforce the idea that there are some systematic differences in their internal composition strategies that lead to systematic differences in how they encode and retain form-related properties across layers.

\section{Different Composition Strategies}
\label{sec:composition_strategy_discussion}
The experimental results strongly indicate that the six LLMs can be categorized into three distinct groups. This pattern emerges consistently across our geometry analysis and probing tasks, suggesting that these differences stem from systematic variations in composition strategies rather than random noise.

The first group, which includes Aya and Gemma2, demonstrates a strong structural alignment between composed representations and original word representations across all layers. These models maintain high precision in geometry experiments, and their word type and length prediction performance remains stable, suggesting that both relevant information are generally well preserved. This implies that these models use a relatively direct and stable composition strategy, where subword embeddings are combined in a way that closely resembles the whole-word embedding throughout all layers. The fact that geometries are isometric to a very large degree, and both form-related and content-related attributes are restored, means the derivation history is implicitly kept, making the input more easily derivable from the output. 

The second group, represented by Falcon and Qwen2.5, follows a different trend. In early layers, their composed representations exhibit good structural similarity with whole-word representations, but this alignment weakens in later layers. The word type semantic information remains relatively stable, but form-related information such as word length disappears in mid-layers and re-emerges towards the end. This suggests that these models initially retain subword structures but shift towards more abstract representations in deeper layers. Rather than maintaining a fixed composition throughout, they seem to undergo a transformation process where subword-based structure gives way to more semantic abstraction. 

The third group, consisting of Llama3 and Llama3.1, exhibits a rapid loss of structural similarity beyond the embedding layer. While the word type prediction results indicate that semantic content is still preserved, form-related features degrade much more quickly than in the other groups. This suggests a more aggressive abstraction process where subword compositions are quickly absorbed into high-level representations, losing their original structural alignment. Unlike the first group, which retains subword traces throughout, these models prioritize semantic fusion over maintaining direct compositional structure.

As discussed in Section \ref{sec:instruct_tuning}, such distinct patterns are already established during pre-training phase. Given the similarities in model architecture and training paradigms across these LLMs, we hypothesize that the main factor leading to this distinction is pre-training data and its data mixture. However, since such information is not fully disclosed\footnote{We include all available information on data mixture in the appendix.} for the models we experimented, drawing a definitive conclusion remains challenging. We hope our work provides insights for future work on exploring different composition strategies.   

\section{Conclusion}
In this work, we examine subword compositionality from the perspective of vector spaces, focusing on three key dimensions: structural similarity, content, and form understanding. Experimental results demonstrate that certain composition operations produce representations that are structurally similar to the original word representations. Additionally, we conducted two probing tasks to analyze content and form information. The results show that content information is consistently preserved across different models and layers, while the preservation of form information exhibits a more variable pattern. The performance of six different LLMs reveals three distinct groups based on their composition strategies. 

\section*{Limitations} 
Our work provides valuable insights into subword composition in LLMs, but several limitations should be noted. First, the size of our dataset (3,432 words) reflects a trade-off between the number of models analyzed and the number of words included. Since different models have varying vocabularies, selecting words (and subwords) that exist across all models required balancing dataset size and model coverage. While carefully selected, the dataset may not fully capture the full range of word structures. Expanding it could offer an even more comprehensive understanding. Additionally, our work focuses on two-subword composition. It would be valuable to extend to compositions with more subwords. Second, our analysis is focused on English, and it remains an open question whether the same composition strategies hold across languages with different morphological properties. Extending this study to other languages would provide a broader perspective on subword composition in LLMs. Third, we have identified three distinct composition strategies, but the underlying reasons for these differences remain to be explored. Factors such as pre-training data and data mixture may play a role, and further investigation could shed light on why LLMs adopt these different composition behaviors.


\section*{Ethical Consideration}
We do not anticipate any risks in the work. In this study, our use of existing artifacts is consistent with their intended purposes. The dataset is under the Creative Commons Attribution-ShareAlike 3.0 Unported License. Falcon\footnote{huggingface.co/tiiuae/falcon-7b-instruct} and Qwen2.5\footnote{huggingface.co/Qwen/Qwen2.5-7B-Instruct} models are under Apache-2.0. Aya-expanse models\footnote{huggingface.co/CohereForAI/aya-expanse-8b} are under cc-by-nc-4.0. Llama3\footnote{https://huggingface.co/meta-llama/Meta-Llama-3-8B-Instruct} and Llama3.1\footnote{https://huggingface.co/meta-llama/Llama-3.1-8B-Instruct} are under the Llama3 and Llama3.1 Community License Agreements respectively. Gemma2 models\footnote{https://huggingface.co/google/gemma-2-9b-it} are under the Gemma Terms of Use.

\section*{Acknowledgement}
We would like to thank all anonymous reviewers for their insightful comments and feedback. This work was supported by DisAI - Improving scientific excellence and creativity in combating disinformation with artificial intelligence and language technologies, a project funded by European Union under the Horizon Europe, GA No. \href{https://doi.org/10.3030/101079164}{101079164}.

\bibliography{acl_latex}

\begin{thebibliography}{42}
\providecommand{\natexlab}[1]{#1}

\bibitem[{Achiam et~al.(2023)Achiam, Adler, Agarwal, Ahmad, Akkaya, Aleman, Almeida, Altenschmidt, Altman, Anadkat et~al.}]{achiam2023gpt}
Josh Achiam, Steven Adler, Sandhini Agarwal, Lama Ahmad, Ilge Akkaya, Florencia~Leoni Aleman, Diogo Almeida, Janko Altenschmidt, Sam Altman, Shyamal Anadkat, et~al. 2023.
\newblock Gpt-4 technical report.
\newblock \emph{arXiv preprint arXiv:2303.08774}.

\bibitem[{Almazrouei et~al.(2023{\natexlab{a}})Almazrouei, Alobeidli, Alshamsi, Cappelli, Cojocaru, Debbah, Goffinet, Heslow, Launay, Malartic, Noune, Pannier, and Penedo}]{falcon40b}
Ebtesam Almazrouei, Hamza Alobeidli, Abdulaziz Alshamsi, Alessandro Cappelli, Ruxandra Cojocaru, Merouane Debbah, Etienne Goffinet, Daniel Heslow, Julien Launay, Quentin Malartic, Badreddine Noune, Baptiste Pannier, and Guilherme Penedo. 2023{\natexlab{a}}.
\newblock {Falcon-40B}: an open large language model with state-of-the-art performance.

\bibitem[{Almazrouei et~al.(2023{\natexlab{b}})Almazrouei, Alobeidli, Alshamsi, Cappelli, Cojocaru, Debbah, Goffinet, Hesslow, Launay, Malartic et~al.}]{almazrouei2023falcon}
Ebtesam Almazrouei, Hamza Alobeidli, Abdulaziz Alshamsi, Alessandro Cappelli, Ruxandra Cojocaru, M{\'e}rouane Debbah, {\'E}tienne Goffinet, Daniel Hesslow, Julien Launay, Quentin Malartic, et~al. 2023{\natexlab{b}}.
\newblock The falcon series of open language models.
\newblock \emph{arXiv preprint arXiv:2311.16867}.

\bibitem[{Batsuren et~al.(2022)Batsuren, Bella, Arora, Martinovic, Gorman, {\v{Z}}abokrtsk{\'y}, Ganbold, Dohnalov{\'a}, {\v{S}}ev{\v{c}}{\'i}kov{\'a}, Pelegrinov{\'a}, Giunchiglia, Cotterell, and Vylomova}]{batsuren-etal-2022-sigmorphon}
Khuyagbaatar Batsuren, G{\'a}bor Bella, Aryaman Arora, Viktor Martinovic, Kyle Gorman, Zden{\v{e}}k {\v{Z}}abokrtsk{\'y}, Amarsanaa Ganbold, {\v{S}}{\'a}rka Dohnalov{\'a}, Magda {\v{S}}ev{\v{c}}{\'i}kov{\'a}, Kate{\v{r}}ina Pelegrinov{\'a}, Fausto Giunchiglia, Ryan Cotterell, and Ekaterina Vylomova. 2022.
\newblock \href {https://doi.org/10.18653/v1/2022.sigmorphon-1.11} {The {SIGMORPHON} 2022 shared task on morpheme segmentation}.
\newblock In \emph{Proceedings of the 19th SIGMORPHON Workshop on Computational Research in Phonetics, Phonology, and Morphology}, pages 103--116, Seattle, Washington. Association for Computational Linguistics.

\bibitem[{Batsuren et~al.(2024)Batsuren, Vylomova, Dankers, Delgerbaatar, Uzan, Pinter, and Bella}]{batsuren2024evaluating}
Khuyagbaatar Batsuren, Ekaterina Vylomova, Verna Dankers, Tsetsuukhei Delgerbaatar, Omri Uzan, Yuval Pinter, and G{\'a}bor Bella. 2024.
\newblock Evaluating subword tokenization: Alien subword composition and oov generalization challenge.
\newblock \emph{arXiv preprint arXiv:2404.13292}.

\bibitem[{Bertolini et~al.(2021)Bertolini, Weeds, Weir, and Peng}]{bertolini-etal-2021-representing}
Lorenzo Bertolini, Julie Weeds, David Weir, and Qiwei Peng. 2021.
\newblock \href {https://doi.org/10.18653/v1/2021.findings-acl.296} {Representing syntax and composition with geometric transformations}.
\newblock In \emph{Findings of the Association for Computational Linguistics: ACL-IJCNLP 2021}, pages 3343--3353, Online. Association for Computational Linguistics.

\bibitem[{Block(1981)}]{Block1981-BLOPAB}
Ned Block. 1981.
\newblock \href {https://doi.org/10.2307/2184371} {Psychologism and behaviorism}.
\newblock \emph{Philosophical Review}, 90(1):5--43.

\bibitem[{Cao et~al.(2023)Cao, Kojima, Matsuo, and Iwasawa}]{cao-etal-2023-unnatural}
Qi~Cao, Takeshi Kojima, Yutaka Matsuo, and Yusuke Iwasawa. 2023.
\newblock \href {https://doi.org/10.18653/v1/2023.emnlp-main.550} {Unnatural error correction: {GPT}-4 can almost perfectly handle unnatural scrambled text}.
\newblock In \emph{Proceedings of the 2023 Conference on Empirical Methods in Natural Language Processing}, pages 8898--8913, Singapore. Association for Computational Linguistics.

\bibitem[{Chai et~al.(2024{\natexlab{a}})Chai, Fang, Peng, and Li}]{chai-etal-2024-tokenization}
Yekun Chai, Yewei Fang, Qiwei Peng, and Xuhong Li. 2024{\natexlab{a}}.
\newblock \href {https://doi.org/10.18653/v1/2024.findings-emnlp.86} {Tokenization falling short: On subword robustness in large language models}.
\newblock In \emph{Findings of the Association for Computational Linguistics: EMNLP 2024}, pages 1582--1599, Miami, Florida, USA. Association for Computational Linguistics.

\bibitem[{Chai et~al.(2024{\natexlab{b}})Chai, Liu, Xiao, Wang, Sun, and Wu}]{chai-etal-2024-autoregressive}
Yekun Chai, Qingyi Liu, Jingwu Xiao, Shuohuan Wang, Yu~Sun, and Hua Wu. 2024{\natexlab{b}}.
\newblock \href {https://doi.org/10.18653/v1/2024.emnlp-main.182} {Autoregressive pre-training on pixels and texts}.
\newblock In \emph{Proceedings of the 2024 Conference on Empirical Methods in Natural Language Processing}, pages 3106--3125, Miami, Florida, USA. Association for Computational Linguistics.

\bibitem[{Conneau et~al.(2018)Conneau, Kruszewski, Lample, Barrault, and Baroni}]{conneau-etal-2018-cram}
Alexis Conneau, German Kruszewski, Guillaume Lample, Lo{\"i}c Barrault, and Marco Baroni. 2018.
\newblock \href {https://doi.org/10.18653/v1/P18-1198} {What you can cram into a single {\$}{\&}!{\#}* vector: Probing sentence embeddings for linguistic properties}.
\newblock In \emph{Proceedings of the 56th Annual Meeting of the Association for Computational Linguistics (Volume 1: Long Papers)}, pages 2126--2136, Melbourne, Australia. Association for Computational Linguistics.

\bibitem[{Dang et~al.(2024)Dang, Singh, D'souza, Ahmadian, Salamanca, Smith, Peppin, Hong, Govindassamy, Zhao, Kublik, Amer, Aryabumi, Campos, Tan, Kocmi, Strub, Grinsztajn, Flet-Berliac, Locatelli, Lin, Talupuru, Venkitesh, Cairuz, Yang, Chung, Ko, Shi, Shukayev, Bae, Piktus, Castagné, Cruz-Salinas, Kim, Crawhall-Stein, Morisot, Roy, Blunsom, Zhang, Gomez, Frosst, Fadaee, Ermis, Üstün, and Hooker}]{dang2024ayaexpansecombiningresearch}
John Dang, Shivalika Singh, Daniel D'souza, Arash Ahmadian, Alejandro Salamanca, Madeline Smith, Aidan Peppin, Sungjin Hong, Manoj Govindassamy, Terrence Zhao, Sandra Kublik, Meor Amer, Viraat Aryabumi, Jon~Ander Campos, Yi-Chern Tan, Tom Kocmi, Florian Strub, Nathan Grinsztajn, Yannis Flet-Berliac, Acyr Locatelli, Hangyu Lin, Dwarak Talupuru, Bharat Venkitesh, David Cairuz, Bowen Yang, Tim Chung, Wei-Yin Ko, Sylvie~Shang Shi, Amir Shukayev, Sammie Bae, Aleksandra Piktus, Roman Castagné, Felipe Cruz-Salinas, Eddie Kim, Lucas Crawhall-Stein, Adrien Morisot, Sudip Roy, Phil Blunsom, Ivan Zhang, Aidan Gomez, Nick Frosst, Marzieh Fadaee, Beyza Ermis, Ahmet Üstün, and Sara Hooker. 2024.
\newblock \href {https://arxiv.org/abs/2412.04261} {Aya expanse: Combining research breakthroughs for a new multilingual frontier}.
\newblock \emph{Preprint}, arXiv:2412.04261.

\bibitem[{Dasgupta et~al.(2018)Dasgupta, Guo, Stuhlm{\"u}ller, Gershman, and Goodman}]{dasgupta2018evaluating}
Ishita Dasgupta, Demi Guo, Andreas Stuhlm{\"u}ller, Samuel~J Gershman, and Noah~D Goodman. 2018.
\newblock Evaluating compositionality in sentence embeddings.
\newblock \emph{arXiv preprint arXiv:1802.04302}.

\bibitem[{Dubey et~al.(2024)Dubey, Jauhri, Pandey, Kadian, Al-Dahle, Letman, Mathur, Schelten, Yang, Fan et~al.}]{dubey2024llama}
Abhimanyu Dubey, Abhinav Jauhri, Abhinav Pandey, Abhishek Kadian, Ahmad Al-Dahle, Aiesha Letman, Akhil Mathur, Alan Schelten, Amy Yang, Angela Fan, et~al. 2024.
\newblock The llama 3 herd of models.
\newblock \emph{arXiv preprint arXiv:2407.21783}.

\bibitem[{Dziri et~al.(2024)Dziri, Lu, Sclar, Li, Jiang, Lin, Welleck, West, Bhagavatula, Le~Bras et~al.}]{dziri2024faith}
Nouha Dziri, Ximing Lu, Melanie Sclar, Xiang~Lorraine Li, Liwei Jiang, Bill~Yuchen Lin, Sean Welleck, Peter West, Chandra Bhagavatula, Ronan Le~Bras, et~al. 2024.
\newblock Faith and fate: Limits of transformers on compositionality.
\newblock \emph{Advances in Neural Information Processing Systems}, 36.

\bibitem[{Ettinger et~al.(2018)Ettinger, Elgohary, Phillips, and Resnik}]{ettinger-etal-2018-assessing}
Allyson Ettinger, Ahmed Elgohary, Colin Phillips, and Philip Resnik. 2018.
\newblock \href {https://aclanthology.org/C18-1152/} {Assessing composition in sentence vector representations}.
\newblock In \emph{Proceedings of the 27th International Conference on Computational Linguistics}, pages 1790--1801, Santa Fe, New Mexico, USA. Association for Computational Linguistics.

\bibitem[{Gee et~al.(2022)Gee, Zugarini, Rigutini, and Torroni}]{gee-etal-2022-fast}
Leonidas Gee, Andrea Zugarini, Leonardo Rigutini, and Paolo Torroni. 2022.
\newblock \href {https://doi.org/10.18653/v1/2022.emnlp-industry.41} {Fast vocabulary transfer for language model compression}.
\newblock In \emph{Proceedings of the 2022 Conference on Empirical Methods in Natural Language Processing: Industry Track}, pages 409--416, Abu Dhabi, UAE. Association for Computational Linguistics.

\bibitem[{Gong et~al.(2017)Gong, Bhat, and Viswanath}]{gong2017geometry}
Hongyu Gong, Suma Bhat, and Pramod Viswanath. 2017.
\newblock Geometry of compositionality.
\newblock In \emph{Proceedings of the AAAI Conference on Artificial Intelligence}, volume~31.

\bibitem[{Hewitt and Manning(2019)}]{hewitt-manning-2019-structural}
John Hewitt and Christopher~D. Manning. 2019.
\newblock \href {https://doi.org/10.18653/v1/N19-1419} {{A} structural probe for finding syntax in word representations}.
\newblock In \emph{Proceedings of the 2019 Conference of the North {A}merican Chapter of the Association for Computational Linguistics: Human Language Technologies, Volume 1 (Long and Short Papers)}, pages 4129--4138, Minneapolis, Minnesota. Association for Computational Linguistics.

\bibitem[{Klafka and Ettinger(2020)}]{klafka-ettinger-2020-spying}
Josef Klafka and Allyson Ettinger. 2020.
\newblock \href {https://doi.org/10.18653/v1/2020.acl-main.434} {Spying on your neighbors: Fine-grained probing of contextual embeddings for information about surrounding words}.
\newblock In \emph{Proceedings of the 58th Annual Meeting of the Association for Computational Linguistics}, pages 4801--4811, Online. Association for Computational Linguistics.

\bibitem[{Kudo(2018)}]{kudo-2018-subword}
Taku Kudo. 2018.
\newblock \href {https://doi.org/10.18653/v1/P18-1007} {Subword regularization: Improving neural network translation models with multiple subword candidates}.
\newblock In \emph{Proceedings of the 56th Annual Meeting of the Association for Computational Linguistics (Volume 1: Long Papers)}, pages 66--75, Melbourne, Australia. Association for Computational Linguistics.

\bibitem[{Li et~al.(2024{\natexlab{a}})Li, Kementchedjhieva, Fierro, and S{\o}gaard}]{li2024vision}
Jiaang Li, Yova Kementchedjhieva, Constanza Fierro, and Anders S{\o}gaard. 2024{\natexlab{a}}.
\newblock Do vision and language models share concepts? a vector space alignment study.
\newblock \emph{Transactions of the Association for Computational Linguistics}, 12:1232--1249.

\bibitem[{Li et~al.(2024{\natexlab{b}})Li, Jiang, Xie, Song, Lian, and Wei}]{li-etal-2024-understanding}
Zhaoyi Li, Gangwei Jiang, Hong Xie, Linqi Song, Defu Lian, and Ying Wei. 2024{\natexlab{b}}.
\newblock \href {https://doi.org/10.18653/v1/2024.findings-acl.576} {Understanding and patching compositional reasoning in {LLM}s}.
\newblock In \emph{Findings of the Association for Computational Linguistics: ACL 2024}, pages 9668--9688, Bangkok, Thailand. Association for Computational Linguistics.

\bibitem[{Lozhkov et~al.(2024)Lozhkov, Li, Allal, Cassano, Lamy-Poirier, Tazi, Tang, Pykhtar, Liu, Wei et~al.}]{lozhkov2024starcoder}
Anton Lozhkov, Raymond Li, Loubna~Ben Allal, Federico Cassano, Joel Lamy-Poirier, Nouamane Tazi, Ao~Tang, Dmytro Pykhtar, Jiawei Liu, Yuxiang Wei, et~al. 2024.
\newblock Starcoder 2 and the stack v2: The next generation.
\newblock \emph{arXiv preprint arXiv:2402.19173}.

\bibitem[{Peng and S{\o}gaard(2024)}]{peng-sogaard-2024-concept}
Qiwei Peng and Anders S{\o}gaard. 2024.
\newblock \href {https://doi.org/10.18653/v1/2024.emnlp-main.315} {Concept space alignment in multilingual {LLM}s}.
\newblock In \emph{Proceedings of the 2024 Conference on Empirical Methods in Natural Language Processing}, pages 5511--5526, Miami, Florida, USA. Association for Computational Linguistics.

\bibitem[{Rust et~al.(2023)Rust, Lotz, Bugliarello, Salesky, de~Lhoneux, and Elliott}]{rust2023language}
Phillip Rust, Jonas~F. Lotz, Emanuele Bugliarello, Elizabeth Salesky, Miryam de~Lhoneux, and Desmond Elliott. 2023.
\newblock \href {https://openreview.net/forum?id=FkSp8VW8RjH} {Language modelling with pixels}.
\newblock In \emph{The Eleventh International Conference on Learning Representations}.

\bibitem[{Sch{\"o}nemann(1966)}]{schonemann1966generalized}
Peter~H Sch{\"o}nemann. 1966.
\newblock A generalized solution of the orthogonal procrustes problem.
\newblock \emph{Psychometrika}, 31(1):1--10.

\bibitem[{Schuster and Nakajima(2012)}]{schuster2012japanese}
Mike Schuster and Kaisuke Nakajima. 2012.
\newblock Japanese and korean voice search.
\newblock In \emph{2012 IEEE international conference on acoustics, speech and signal processing (ICASSP)}, pages 5149--5152. IEEE.

\bibitem[{Sennrich et~al.(2016)Sennrich, Haddow, and Birch}]{sennrich-etal-2016-neural}
Rico Sennrich, Barry Haddow, and Alexandra Birch. 2016.
\newblock \href {https://doi.org/10.18653/v1/P16-1162} {Neural machine translation of rare words with subword units}.
\newblock In \emph{Proceedings of the 54th Annual Meeting of the Association for Computational Linguistics (Volume 1: Long Papers)}, pages 1715--1725, Berlin, Germany. Association for Computational Linguistics.

\bibitem[{Shani et~al.(2023)Shani, Vreeken, and Shahaf}]{shani-etal-2023-towards}
Chen Shani, Jilles Vreeken, and Dafna Shahaf. 2023.
\newblock \href {https://doi.org/10.18653/v1/2023.findings-emnlp.877} {Towards concept-aware large language models}.
\newblock In \emph{Findings of the Association for Computational Linguistics: EMNLP 2023}, pages 13158--13170, Singapore. Association for Computational Linguistics.

\bibitem[{Tai et~al.(2024)Tai, Liao, Suglia, and Vergari}]{tai-etal-2024-pixar}
Yintao Tai, Xiyang Liao, Alessandro Suglia, and Antonio Vergari. 2024.
\newblock \href {https://doi.org/10.18653/v1/2024.findings-acl.874} {{PIXAR}: Auto-regressive language modeling in pixel space}.
\newblock In \emph{Findings of the Association for Computational Linguistics: ACL 2024}, pages 14673--14695, Bangkok, Thailand. Association for Computational Linguistics.

\bibitem[{Team et~al.(2023)Team, Anil, Borgeaud, Alayrac, Yu, Soricut, Schalkwyk, Dai, Hauth, Millican et~al.}]{team2023gemini}
Gemini Team, Rohan Anil, Sebastian Borgeaud, Jean-Baptiste Alayrac, Jiahui Yu, Radu Soricut, Johan Schalkwyk, Andrew~M Dai, Anja Hauth, Katie Millican, et~al. 2023.
\newblock Gemini: a family of highly capable multimodal models.
\newblock \emph{arXiv preprint arXiv:2312.11805}.

\bibitem[{Team(2024{\natexlab{a}})}]{gemma_2024}
Gemma Team. 2024{\natexlab{a}}.
\newblock \href {https://doi.org/10.34740/KAGGLE/M/3301} {Gemma}.

\bibitem[{Team(2024{\natexlab{b}})}]{team2024qwen2}
Qwen Team. 2024{\natexlab{b}}.
\newblock Qwen2. 5 technical report.
\newblock \emph{arXiv preprint arXiv:2412.15115}.

\bibitem[{Touvron et~al.(2023)Touvron, Martin, Stone, Albert, Almahairi, Babaei, Bashlykov, Batra, Bhargava, Bhosale et~al.}]{touvron2023llama}
Hugo Touvron, Louis Martin, Kevin Stone, Peter Albert, Amjad Almahairi, Yasmine Babaei, Nikolay Bashlykov, Soumya Batra, Prajjwal Bhargava, Shruti Bhosale, et~al. 2023.
\newblock Llama 2: Open foundation and fine-tuned chat models.
\newblock \emph{arXiv preprint arXiv:2307.09288}.

\bibitem[{Wang et~al.(2020)Wang, Cho, and Gu}]{wang2020neural}
Changhan Wang, Kyunghyun Cho, and Jiatao Gu. 2020.
\newblock Neural machine translation with byte-level subwords.
\newblock In \emph{Proceedings of the AAAI conference on artificial intelligence}, volume~34, pages 9154--9160.

\bibitem[{Wang et~al.(2024{\natexlab{a}})Wang, Li, Jiang, Ding, Jiang, Liang, and Yang}]{wang2024tokenization}
Dixuan Wang, Yanda Li, Junyuan Jiang, Zepeng Ding, Guochao Jiang, Jiaqing Liang, and Deqing Yang. 2024{\natexlab{a}}.
\newblock Tokenization matters! degrading large language models through challenging their tokenization.
\newblock \emph{arXiv preprint arXiv:2405.17067}.

\bibitem[{Wang et~al.(2024{\natexlab{b}})Wang, Wei, Choi, and Ren}]{wang-etal-2024-llms}
Siyuan Wang, Zhongyu Wei, Yejin Choi, and Xiang Ren. 2024{\natexlab{b}}.
\newblock \href {https://doi.org/10.18653/v1/2024.acl-long.406} {Can {LLM}s reason with rules? logic scaffolding for stress-testing and improving {LLM}s}.
\newblock In \emph{Proceedings of the 62nd Annual Meeting of the Association for Computational Linguistics (Volume 1: Long Papers)}, pages 7523--7543, Bangkok, Thailand. Association for Computational Linguistics.

\bibitem[{Wu et~al.(2024)Wu, Lei, Yates, and Monz}]{wu-etal-2024-representational}
Di~Wu, Yibin Lei, Andrew Yates, and Christof Monz. 2024.
\newblock \href {https://doi.org/10.18653/v1/2024.findings-emnlp.823} {Representational isomorphism and alignment of multilingual large language models}.
\newblock In \emph{Findings of the Association for Computational Linguistics: EMNLP 2024}, pages 14074--14085, Miami, Florida, USA. Association for Computational Linguistics.

\bibitem[{Xu et~al.(2024)Xu, Zhang, Zhang, Qian, and Huang}]{xu2024tip}
Ningyu Xu, Qi~Zhang, Menghan Zhang, Peng Qian, and Xuanjing Huang. 2024.
\newblock On the tip of the tongue: Analyzing conceptual representation in large language models with reverse-dictionary probe.
\newblock \emph{arXiv preprint arXiv:2402.14404}.

\bibitem[{Xu et~al.(2023)Xu, Guo, and Cristianini}]{xu2023compositionality}
Zhaozhen Xu, Zhijin Guo, and Nello Cristianini. 2023.
\newblock On compositionality in data embedding.
\newblock In \emph{International Symposium on Intelligent Data Analysis}, pages 484--496. Springer.

\bibitem[{Yu and Ettinger(2020)}]{yu-ettinger-2020-assessing}
Lang Yu and Allyson Ettinger. 2020.
\newblock \href {https://doi.org/10.18653/v1/2020.emnlp-main.397} {Assessing phrasal representation and composition in transformers}.
\newblock In \emph{Proceedings of the 2020 Conference on Empirical Methods in Natural Language Processing (EMNLP)}, pages 4896--4907, Online. Association for Computational Linguistics.

\end{thebibliography}

\appendix

\section{Data Mixture for Different LLMs}
\paragraph{Llama3 and Llama3.1} As revealed in \citet{dubey2024llama}, the final data mix for Llama3 pretraining contains roughly 50\% of tokens corresponding to general knowledge, 25\% of mathematical and reasoning tokens, 17\% code tokens, and 8\% multilingual tokens.

\begin{figure}[!ht]
    \centering
    \includegraphics[width=1\columnwidth]{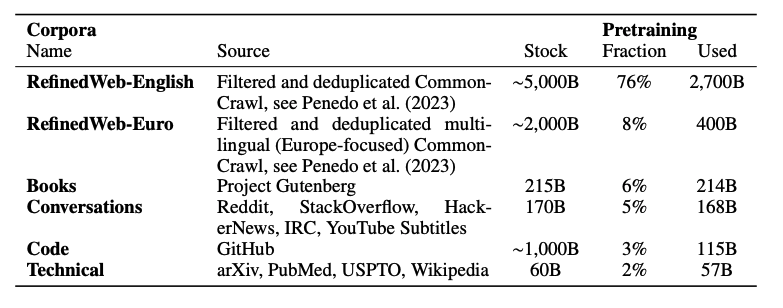}
    \caption{The figure taken from \citet{almazrouei2023falcon} that illustrates pre-training data mixture in Falcon models.}
    \label{fig:falcon_data_mix}
\end{figure}

\paragraph{Falcon} The pre-training data mixture for Falcon is summarized in Figure \ref{fig:falcon_data_mix}.

\paragraph{Gemma2} Gemma 2 models (9B) are pre-trained on 8 trillion tokens. These
tokens come from a variety of data sources, including web documents, code, and science articles. However, exact proportions of these data types are not disclosed. Instead, it is noted that the final data mixture was determined through ablations similar to the approach in Gemini 1.0 \citep{team2023gemini}.

\paragraph{Aya-expanse} The details of pre-training data mixture is not mentioned or discussed in \citet{dang2024ayaexpansecombiningresearch}.

\paragraph{Qwen2.5} The fraction of data mixture for Qwen2.5 models are not revealed. \citet{team2024qwen2} mentions that they employ Qwen2-Instruct models to optimize the pre-training data distribution across different domains and results in a pre-training data of 18 trillion tokens. 

\paragraph{Tokenizers} Tokenizers of these different LLMs all adopt the BPE algorithm and give quite similar results for tokenization. This is also why we choose these six models as they give the highest overlap in words, ensuring enough data to experiment with. 



\end{document}